\documentclass[10pt,twocolumn,letterpaper]{article}

\usepackage{iccv}
\usepackage{times}
\usepackage{epsfig}
\usepackage{graphicx}
\usepackage{amsmath}
\usepackage{amssymb}

\iccvfinalcopy % *** Uncomment this line for the final submission

\def\iccvPaperID{****} % *** Enter the ICCV Paper ID here

\ificcvfinal\fi
\usepackage{graphicx}
\usepackage{amsmath}
\usepackage{amssymb}
\usepackage{booktabs}

\usepackage{tabularx} % in the preamble
\usepackage{times}
\usepackage{epsfig}
\usepackage{graphicx}
\usepackage{amsmath}
\usepackage{amssymb}
\usepackage{siunitx}
\usepackage{mycommands}
\usepackage{color}
\usepackage{pifont}

\usepackage{tabularx}
\usepackage{xpatch}

\usepackage{floatrow}

\usepackage[capitalize]{cleveref}
\crefname{section}{Sec.}{Secs.}
\Crefname{section}{Section}{Sections}
\Crefname{table}{Table}{Tables}
\crefname{table}{Tab.}{Tabs.}

\def\iccvPaperID{9283} % *** Enter the iccv Paper ID here

\usepackage{amsmath}

\begin{document}

%%%%%%%%% TITLE
% \title{Constrained Nearest Neighbors: Boosting local feature-based retrieval for visual localization}

\title{Yes, we CANN: Constrained Approximate Nearest Neighbors for local feature-based visual localization}

% Previous ones:
% \title{Constrained Nearest Neighbors: Boosting performance of local features for image retrieval in visual localization}

\author{
%\phantom{000000000} 
Dror Aiger \hspace{0.3in}
André Araujo    \hspace{0.3in}
Simon Lynen    \vspace{0.1in}\\
Google Research \\
{\tt\small \{aigerd,andrearaujo,slynen\}@google.com}
}

\maketitle

%%%%%%%%% ABSTRACT
\begin{abstract}

Large-scale visual localization systems continue to rely on 3D point clouds built from image collections using structure-from-motion. While the 3D points in these models are represented using local image features, directly matching a query image's local features against the point cloud is challenging due to the scale of the nearest-neighbor search problem. Many recent approaches to visual localization have thus proposed a hybrid method, where first a global (per image) embedding is used to retrieve a small subset of database images, and local features of the query are matched only against those. It seems to have become common belief that global embeddings are critical for said image-retrieval in visual localization, despite the significant downside of having to compute two feature types for each query image. In this paper, we take a step back from this assumption and propose Constrained Approximate Nearest Neighbors (CANN), a joint solution of k-nearest-neighbors across both the geometry and appearance space using only local features. We first derive the theoretical foundation for k-nearest-neighbor retrieval across multiple metrics and then showcase how CANN improves visual localization. Our experiments on public localization benchmarks demonstrate that our method significantly outperforms both state-of-the-art global feature-based retrieval and approaches using local feature aggregation schemes. Moreover, it is an order of magnitude faster in both index and query time than feature aggregation schemes for these datasets. Code:  \url{https://github.com/google-research/google-research/tree/master/cann}

%Code is released\footnote{https://github.com/google-research/google-research/tree/master/cann\label{fnlabel}}

\end{abstract}

% \dror{given that we have HOW+ASMK similar to HOW+CANN and R2D2+CANN for global sfm (not better as we had in EWB when we tuned for it) in both baidu and aachen (tuned on baidu 300 queries), i would accept andre's suggestions to change the story a bit and submit. in the worst case it will be rejected and we will get comments for ICCV. i'm finalizing the plots and then will get back to the text.

% - CANN is a replacement for ASMK which is much simpler and faster in both indexing and query, working purely on $L_p$ distances in descriptor space, avoiding kmeans and aggregation.

% - CANN+R2D2 (the same localization features existing anyway) outperforms all SOTA global features for both metrics (EWB and global sfm) considerably and that is all published work for localization so far.

% - in a very recent unpublished work (private communication with the authors of []), after we started the research, there are results with HOW+ASMK and FIRE+ASMK. we match HOW results and outperform FIRE.  we can put how and fire only in the sup. mat. or in the main plots.

% - in the sup. mat we can add more datasets, i believe they will be they same.

% what do you think?}

%%%%%%%%% BODY TEXT
\section{Introduction}
\label{sec:intro}

% Testing a citation: \cite{noh2017large}.
% This is an example of a TODO: \todo{write this intro}
% \andre{Andre's text example}
% \dror{Dror's text example}
% \simon{Simon's text example}
% \manuel{Manuel's text example}

\begin{figure}[t]
\begin{center}
   \includegraphics[width=1.0\linewidth]{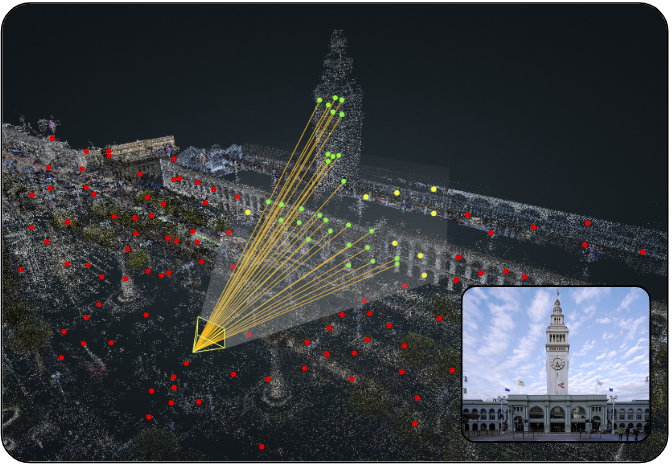}
\end{center}
\vspace{-10pt}
   \caption{The proposed Constrained Approximate Nearest Neighbor algorithm allows to find the best subset of 3D points that are both close to query features in appearance space and that are consistently seen by the same camera, leading to high overlap with the initially unknown query camera pose (shaded area).
   Jointly solving for these two metrics in a single search algorithm is a long-known open question in the community and CANN provides to the best of our knowledge the first practical solution.
   Red points in the figure show neighbors retrieved by an unconstrained search using the features from the query image (bottom right). Using CANN it's more likely to retrieve points that are inliers to geometric verification (green) and less likely to fetch unrelated outlier points (yellow).
   }
\label{fig:teaser}
\end{figure}

\noindent In this paper we focus on the problem of image retrieval for visual localization.
Modern visual localization approaches are predominantly based on 3D point clouds that represent the geometry and appearance of large scale scenes \cite{GeospatialAPI,GeospatialAnchors,sarlin2021back,lynen2020large}.
These 3D points are estimated from image collections using Structure-from-Motion (SfM), where each 3D point has an associated descriptor derived from pixels.

To localize a query image against such 3D models, a set of local features is extracted from it and 2D-3D correspondences are estimated based on descriptor similarity. 
In practice, this data association problem suffers from various challenges: visual aliasing, scene change, noise, etc.
Because the final localization solution is computed using geometric inference from these 2D-3D correspondences, not finding enough correct matches can lead the entire localization process to fail.

Simply establishing many more matches per query keypoint (red points in \figref{fig:teaser}) however causes long runtime in geometric verification \cite{aiger2021efficient}.
It is thus important to find a small 2D-3D set which has high probability to contain ``good" matches (yellow/green points in \figref{fig:teaser}): In fact we know that the 3D points of “good” matches should all lie inside one (unknown) camera frustum which is the one of the query image (shaded area in \figref{fig:teaser}).

There exist several approximations to this problem, ranging from clustering nearest-neighbor matches in the 3D model's covisibility graph \cite{sarlin2018leveraging} to using image retrieval methods to obtain a small set of candidate images for which local features are matched subsequently \cite{sarlin2019from}.
The latter approach, leveraging recent advances in global (per image) embeddings, has gained substantial traction recently \cite{humenberger2022investigating,sarlin2018leveraging,sarlin2019from,berton2022deep}, to a degree that it appears the community has abandoned the idea of finding a solution that jointly solves for appearance and geometry using local features only.
For example, the benchmark we evaluate on \cite{humenberger2022investigating} didn't even consider local feature based retrieval approach at publication time.

We don't consider the case of using local features closed and therefore propose an approach to obtain matches that are close in appearance space while obtaining geometric consistency at the same time -- which is a long-known open question in the community.

\noindent\textbf{Contributions.}
In this paper we make three contributions:

\textbf{(1)} Our first and main contribution is a new method, referred to as Constrained Approximate Nearest Neighbors (CANN), that efficiently obtains a high quality, small set of 2D-3D correspondences.
CANN performs nearest neighbor search in descriptor space in a constrained manner, so that matches are compact in 3D space.
We provide both a brute-force solution as well as an efficient implementation and associated complexity analysis of this \emph{colored} nearest neighbor search algorithm.

\textbf{(2)} Our second contribution is to make the connection of colored nearest neighbor search to the problem space of image retrieval and localization, proposing a metric to rank cameras, which can serve as a way to evaluate future work in this area.

\textbf{(3)} Lastly we provide an extensive evaluation of both global and local feature based methods on four large scale datasets from \cite{humenberger2022investigating}: ``Baidu-Mall",``Gangnam Station",``RobotCar Seasons" and ``Aachen Day-Night v1.1".
We demonstrate that local feature based methods are not only competitive, but in fact strongly outperform global embedding based approaches; which goes contrary to the trend in the community.
We hope to provide new impulse to techniques that aim for jointly searching in appearance and geometry space, which is more efficient and elegant than previously proposed two-step approaches.

\section{Related Work}
\label{sec:rw}

\noindent\textbf{Visual Localization using local features without image retrieval:} 
A large body of work in visual localization \cite{sattler2011fast,sattler2016efficient,zeisl2015camera,sattler2012improving,li2010location,li2012worldwide,sattler2017large,lynen2020large,arth2009wide,svarm2016city} is based on sparse 3D point clouds built from image collections using Structure-from-Motion (SfM).
These methods directly establish 2D-3D matches between local features from the query image and the descriptors associated with 3D points in the model.
As mentioned before, these matches often contain many outliers and thus directly feeding them to geometric verification is typically impractical\cite{aiger2021efficient}.
Therefore several post-filtering techniques have been proposed, such as clustering in the SfM covisibility graph \cite{sattler2012improving,sattler2016efficient} or applying voting in the space of camera poses \cite{zeisl2015camera,lynen2020large}.
Remaining 2D-3D matches typically have a sufficiently low fraction of outliers, so that they can be efficiently processed by geometric verification, using minimal pose solvers \cite{kneip2011novel,sweeney2015efficient} in a RANSAC \cite{Fischler1981} scheme.

\noindent\textbf{Visual Localization using local features for retrieval and 2D-3D matching:}
Image retrieval approaches promise to both reduce the cost of matching against features in the SfM model and achieving high quality matches by limiting the search to only a subset of the model\cite{pion2020benchmarking}.
Such approaches either remove features that don't belong to top-ranking images or perform an additional matching step to top-ranking images before running geometry verification using the obtained local feature matches.
Our proposed algorithm provides an alternative to these two-step filtering approaches, by directly optimizing for compactness of nearest neighbor matches in the covisibility graph or 3D space \emph{during} the search.

\noindent\textbf{Visual Localization using global features for retrieval and local features for 2D-3D matching:}
Image retrieval using local features however has most recently lost attention from the community and instead global features (\eg, DELG-GLDv2 \cite{cao2020unifying} and AP-GeM \cite{revaud2019learning}) have dominated benchmarks \cite{humenberger2022investigating}.
While using global features offers significant speedups due to the much smaller database size, the full-image embeddings are not appropriate for high quality localization due to their global nature \cite{humenberger2022investigating}.
In order to obtain an accurate localization result, some approaches \cite{sarlin2019from, sarlin2018leveraging} compute additionally local features, which are matched only between the query image and top-ranking images from the database.
While there are attempts to concurrently compute local and global features to reduce cost/latency \cite{cao2020unifying}, the accuracy of the local feature keypoints remain inferior to approaches that compute dedicated local features \cite{sarlin2021back}.

\noindent\textbf{Local feature-based image retrieval techniques:}
Despite the image retrieval community's recent focus on global features, local feature-based retrieval has a long history, with well-established methods \cite{Sivic2003,Philbin07,jegou2012aggregating,tolias2015image,noh2017large}.
Among these, the most relevant method today is the Aggregated Selective
Match Kernels (ASMK), which continues to be explored recently in conjunction with deep-learned local features \cite{teichmann2019detect,tolias2020learning,weinzaepfel2022learning}.
ASMK (like VLAD \cite{arandjelovic2013all}) performs local descriptor aggregation and essentially produces high-dimensional global image representations, which are however sparse and can be searched efficiently.
In contrast, our method operates directly on local descriptor space and avoids aggregation, which makes it more suitable to match against partial views and unique details that do not get lost in aggregation.

\noindent\textbf{Approximate nearest neighbor methods:}
Another related field is the area of proximity problems in high dimensional spaces with its many applications in computer vision \cite{jegou2010product,babenko2014inverted,datar2004locality,aiger2013random} (to name a few).
The most common of this kind is nearest neighbor search, where given a set $P$ of $n$ points in a high-dimensional space $R^d$ we wish to find the point(s) in $P$ closest to a query point $q$.
Extensive research on this problem has led to a variety of interesting solutions, both exact and approximate \cite{Goodman2004}.
In many use cases, indexed points in the ``database" are equipped with additional attributes, such vector-valued attributes or simple scalars, such as an ID (``color") that indicates a grouping of points.

The term Constrained Approximate Nearest Neighbors that we propose in this paper refers to a way to apply nearest neighbors in one space given constraints in the space of these attributes.
The simplest such case is ``colored nearest neighbor search":
each point in $P$ is assigned with an ID and for a given query point $q$ (with or without colors), we want to use the IDs of points in $P$ as constraints during the search.
A simple example, which is the use case in this work, is to return nearest neighbors for all query points, provided that all of the neighbors have the same ID.
The optimal result are those points in $P$ that all have the same ID and optimize some metric, such as the sum of distances to the query points.

Colored range searching and nearest neighbors (also known as “categorical range searching”, or “generalized range searching”) have been extensively studied in computational geometry since the 1990s \cite{janardan1993generalized,gupta1996algorithms,gupta1997technique}.
The colored versions of nearest neighbor (or range searching) problems tend to be harder than their uncolored counterparts and several different problems and solutions were proposed, see e.g. \cite{Gupta2018}.
To the best of our knowledge, no previous problem and solution fits into the requirement that we need in this work and the Constrained Approximate Nearest Neighbor problem we address here is new.

\section{Method}
\label{sec:method}

\begin{figure}[ht]
\begin{center}
   \includegraphics[width=1.0\linewidth]{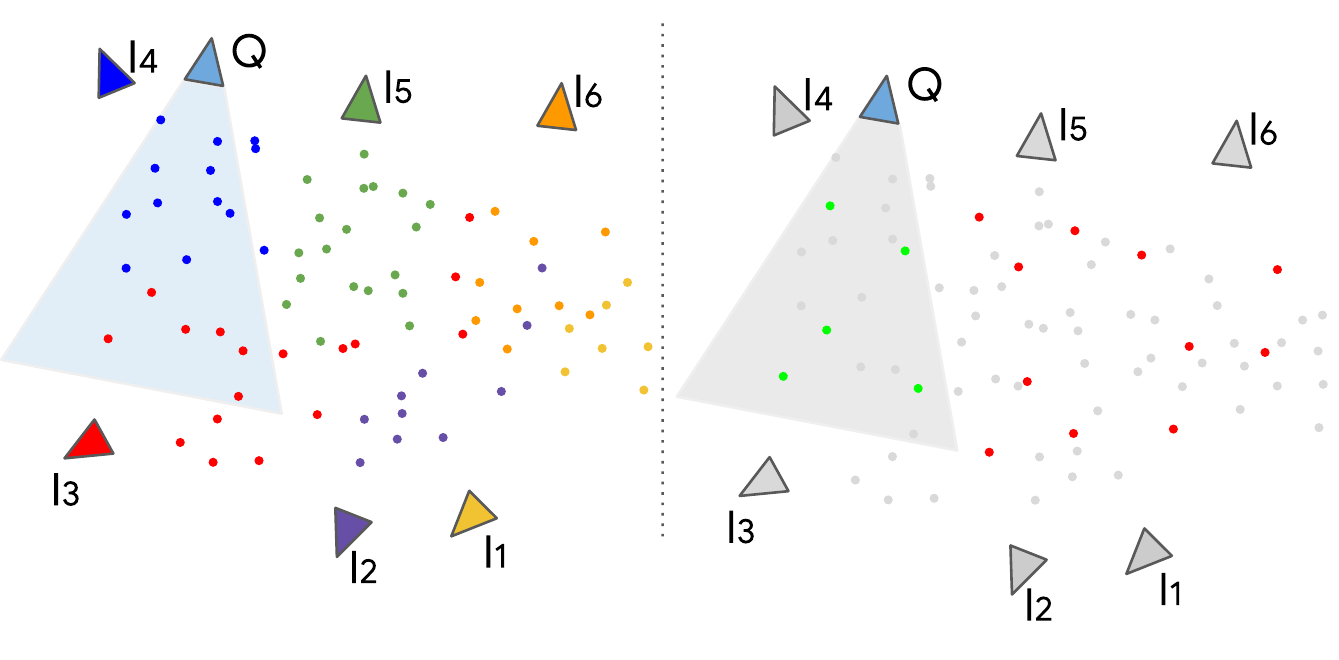}
\end{center}
   \caption{A visual depiction of CANN: the image on the left shows 3D points colored by the camera from which they were reconstructed. CANN leverages this information to retrieve feature matches that are consistently seen in the same camera. This contrasts with prior art (on the right), where unconstrainted feature matching returns many unrelated outlier matches (red), which then need to be filtered out subsequently by geometric verification to obtain inlier matches (green).
   }
\label{fig:cann}
\end{figure}

\subsection{Ranking Images for Visual Localization using Constrained Approximate Nearest Neighbors}

We first propose a natural metric to rank cameras and then show that this ranking can be efficiently computed during the feature-matching stage instead of requiring post processing.
For simplicity of presentation we consider the case of a single optimal camera/image from the index.
This is without loss of generality, since in practice, we may use $k$-best cameras or simply weight matches by the rank of each image.

\textbf{The metric:}
We are given a large $d$-dimensional space containing local feature descriptors extracted from all images in a large collection.
Denote $I=\{0,1,2,\ldots\}$ the set of image IDs in that collection.
We assign each local descriptor the ID of the image, $i \in I$, it was computed from, so we obtain the set $P$ of ``ID-colored" points (see colors in \figref{fig:cann} on the left).
Then, at query time, for a query image with a set of features $Q=\{q_j\}$ extracted from it, let $d_{ij}=d(q_j,NN_i(q_j))/R$ be the (normalized) Euclidean distance in descriptor space between the feature $q_j$ to its nearest neighbor descriptor in image $i$. $R$ is some fixed maximum distance that we use for normalization such that $d_{ij} \in [0,1]$. We then compute a score for each image $i$ in the dataset

\begin{figure}[t]
    \centering
    \includegraphics[width=8.0cm]{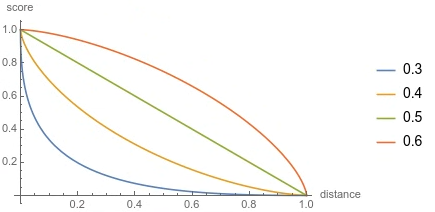}
    \caption{Our score for $R=1$ and various $p$ different values in Equation~\ref{alg:query-r}. $p$ is a parameter of our metric that we tune upfront and is used to compute $s_i$ for all $d_{i,j}$.}
    \label{fig:score}
    \vspace{-10pt}
\end{figure}

\setlength{\belowdisplayskip}{-3pt} \setlength{\belowdisplayshortskip}{3pt}
\setlength{\abovedisplayskip}{-5pt} \setlength{\abovedisplayshortskip}{-5pt}
\begin{align}
s_i=\sum_{j}(1.0-d_{ij}^{\frac{p}{1-p}})^{\frac{1-p}{p}}
\label{eq:metric}
\vspace{-10pt}
\end{align}

\noindent and use it to rank all images with respect to the query image features $q_j \in Q$. To obtain this per-image compact set of descriptors from the set of all indexed descriptors $P$ (with their ``ID-color"), we have to develop an efficient colored version of nearest neighbors. Such algorithm obtains the nearest neighbor of each $q_j$ for all colors at once, provided that its distance is at most $R$. We observe that depending on a tuned parameter $p$, we can crop the distances at $R$ such that all distances larger than $R$ have score at most some very small value (say $10^{-6}$). This allows to get good bound on the runtime of the search for $NN$. Figure \ref{fig:score} shows our metric.

\subsection{Preliminaries} To explain the proposed Constrained Approximate Nearest Neighbors algorithm we refer to standard tools like Approximate Nearest Neighbors (ANN) and Approximate Range searching (RS) and make the (common) assumption that there is a maximum distance, $R$, known at local descriptor indexing time. We also assume that randomization is allowed, i.e. all results are correct with (arbitrary) high probability. Details on the exact definitions of ANN and RS for the case of bounded distance can be found in~\cite{aiger2013}.
We can assume (for simplicity of presentation) that ANN and RS data structures can be created in $O(C_I(d,c)*n)$ and a point query takes $O(C_q(d,c)+k)$ time, $C_q(d,c)$ is a constant depending on the dimension $d$ and the approximation factor, $c$ and $k$ is the maximum number of items it returns. For image retrieval, this runtime is multiplied by the number of features in the image, $|Q|$.

\paragraph{Colored Nearest Neighbors vs Colored Range Searching}As can be seen from Equation~\ref{eq:metric}, we need a colored NN data structure to compute the scores for all relevant images given one query point $q_j$. Such algorithm returns for each $q_j$ the set of $1$-NN in all cameras within radius $R$.
We see from the metric that cameras without such neighbor don't contribute to the sum, so we want as many neighbors with as low Euclidean distance from the query as possible.
We are not aware of any efficient algorithm to perform this
% with a reasonable constant factor approximation in time 
with a better time complexity than a brute force method using $|I|$ separate NN structures (See Section~\ref{CANN-RS}).
Fortunately, we can reduce this colored NN problem to a fixed $R$ colored range searching which can be implemented efficiently.
A reduction from the fixed radius decision problem: ``is there a point within distance $R$ from the query" to the approximate NN is well known from LSH~\cite{DBLP:conf/stoc/IndykM98} using a form of binary search over several $R$'s. While this approach isn't directly applicable for colored searches, we can use similar ideas as outlined in the following section.

\subsection{Colored Range Searching}
In this section we explain the colored nearest neighbor search for computing the scores in Eq.~\eqref{eq:metric}. While there are multiple versions of this problem, we're specifically interested in \emph{colored range reporting:} For a set of colored points in $R^d$, report all the distinct colors in the query range.
Even with approximations, this problem is computationally hard with $O(C_q(d,c)+|I|)$~\cite{Gupta2018,chan2020rs} as lower bound on the runtime.
For a large set of images this bound can be very high, yet in practice it can be solved quite efficiently by introducing the threshold distance $R$.
The most recent work~\cite{chan2020rs} on this exact problem shows that the problem is already hard for low dimensional spaces, even with integer coordinates and considering only orthogonal queries (an axis-aligned box vs. a sphere).
For a set of $n$ colored points in three dimensions, the authors of \cite{chan2020rs} describe a randomized data structure with $O(n*polylog(n))$ space that can report the distinct colors within an axis-aligned box using $O(k*polyloglog(n))$ time, with $k$ as number of distinct colors in the range, assuming that coordinates are in $\{1, . . . , n\}$.
In this paper we show that with $R$ known at index time and allowing for approximation, we can develop a more efficient data structure for the colored range search that allows us to efficiently compute the 1-NN across all images at once. Besides being essential for solving the Constrained Nearest Neighbors problem we believe that this data-structure is interesting on its own and beyond the problem of image localization.

\subsection{A brute force method using ANN}
\label{CANN-RS}
There exist two straightforward algorithms for colored range searching: First build $|I|$ separate regular nearest neighbor structures, one for each color in $O(C_q(d,c)*|P_{I}|*|I|)$ indexing time, with $|P_{I}|$ as the color $I$ in $P$.
Then call them sequentially for each query point $q_j$ with cost $O(C_q(d,c) \times |I|)$. This is independent of $R$ and thus much worse than the above lower bound. The advantage is that the runtime of this version is asymptotically independent of the threshold radius $R$.

The second simple algorithm, that we call CANN-RS, is applicable for small thresholds $R$ using Randomized Range Searching~\cite{aiger2013}:
We index points into a RS data-structure for radius $R$ and then for each query feature, enumerate all neighbors within $R$, keeping a tally of neighbors per image $I$.
Because we only retrieve a small subset of points within radius $R$ we only obtain a few colors (cameras or images) with number of features, much less than $|I|$.
This approach has runtime $O(C_q(d,c) + k)$, here, $k$ is the expected number of neighbors in each such range query over all images. The drawback is that for each query feature we must enumerate \emph{all indexed points in the range R} where most of them do not realize any nearest neighbor for any color. This number ($k$ above) can still be quite large.

\subsection{Efficient algorithm using Random Grids}
\label{sec:CANN-RG}

To implement an efficient variant of the algorithm (CANN-RG), we leverage Random Grids~\cite{aiger2013random,aiger2014reporting}, an efficient $c$-approximate nearest neighbor algorithm based on applying randomized rotations and shifts to high-dimensional vectors prior to hashing them to a set of keys.
We extend the Random Grids to support colored range searching. We show that our algorithm avoids the enumeration of all points in the range $R$ (as in CANN-RS) and doesn't require distance computation in descriptor space which can take most of the time in practice due to high dimensional feature descriptors.

Our algorithm works as follows:
For each query point $q_j$ CANN-RG should report all colors in the range $R$ from $q_i$ approximately by factor $c>1$, i.e. any color that has a feature at distance at most $R$ is reported with high probability and any reported color (image) has a feature at distance at most $cR$.
The points are indexed using Algorithm~\ref{alg:indexing-r}, where we store a set of distinct integers using hash-sets and use hashing to create a key for each non-empty grid cell in a high dimensional space following \cite{aiger2013random,aiger2014reporting}.
At query time we retrieve points from the grid indices using Algorithm~\ref{alg:query-r}.

Note that since we're only interested in the color of points within range $R$, the index only holds point colors not point coordinates and the query results similarly only comprise colors without exact distances.

\begin{algorithm}
\SetAlgoLined
\DontPrintSemicolon
\SetKwFunction{FnIndexColors}{IndexColors}
\SetKwProg{Fn}{Function}{:}{}
\KwData{$P$: $d$-points, $\{p_i\}$ with colors $color(p_i)$ for each $p_i \in P$, $R>0$: Range, $c>1$: Approximation factor}
\KwResult{Index $RG$ for the colors $color(p)$ such that for a given query point $q$, all colors at distance at most $R$ from $q$ are reported quickly.}
\Fn{\FnIndexColors{$P$, $R$, $c$}}{
    Impose a grid of cell size $w = R*c/\sqrt{d}$ on $P$\;
    Create $L$ such grids for $L$ randomly rotated and translated versions of $P$\;
    /* $L$ is determined from the analysis below */\;
    \For{all $p_i \in P$}{
       Add $color(p_i)$ to a distinct set of colors hashed in each of $L$ cells the transformed $p_i$ falls into\;
    }
    /* Each cell now contains a set of distinct colors of all images that have point inside it */\;
    /* NOTE: We do not store the coordinates of the points, just their color (16 bits per color) */\;
    \Return{The set of hashed cells with their lists of colors}\;
}
\caption{Efficient colored Range Searching indexing (CANN-RG-INDEX-R) }
\label{alg:indexing-r}
\end{algorithm}
\vspace{-10pt}

\setlength{\textfloatsep}{0pt}% Remove \textfloatsep
\begin{algorithm}
\SetAlgoLined
\DontPrintSemicolon
\SetKwFunction{FnQueryPoint}{QueryPoint}
\SetKwProg{Fn}{Function}{:}{}
\KwData{$RG$: A Random Grid index for colors (Algorithm \ref{alg:indexing-r}), $q$: query $d$-point}
\KwResult{A set of colors (images) that have a point at distance at most $R$ from $q$}
\Fn{\FnQueryPoint{$RG$,$q$}}{
     Create an empty set distinct colors, $S$\;
     \For{all grids $g \in RG$}{
      Rotate and translate $q$ according to $g$ to obtain $q_t$\;
      Retrieve the set of colors, $colors(g)$, from the grid cell in $g$ that $q_t$ falls into\;
      Insert all colors in $colors(g)$ into $S$\;
     }
     /* $S$ now contains a set of distinct colors that have a point at distance at most $cR$ from $q$ */
\Return{S}\;
}
\caption{Query an arbitrary point in the index built for a given $R$ (CANN-RG-QUERY-R) }
\label{alg:query-r}
\end{algorithm}
\vspace{-10pt}

\paragraph{Analysis}
In this section we analyze indexing and query algorithms of CANN-RG.
First we make concrete the constants $C_I(d,c)$ and $C_q(d,c)$ which appear in all Random Grids implementations: For the grid cell of size $l*c/\sqrt{d}$, a random vector of length $l$ in $R^d$ will be captured in a given cell with probability at least $e^{-\sqrt{d}/w}=e^{-d/c}$~\cite{aiger2014reporting}.
We thus need $L=e^{d/c}$ random grids in Algorithm~\ref{alg:indexing-r} to ensure that, if there is a point in $P$ at distance at most $R$ from $q$, its color will be found in at least one of the grid cells with constant probability.
On the other hand, any color of a point found in a grid cell that also contains $q_t$ (the rotated and translated version of $q$ for that grid) is at distance at most $cR$ from $q$ due to the size of the grid cells.

Because we do not care about the coordinates of indexed points, we only store each color at most once per grid cell. Therefore the data structure build time  $(C_I(d,c)*|P|)=O(e^{d/c}*|P|)$ and storage  $O(e^{d/c}*|P|)$ are linear in $|P|$.
For each query point $q$, we retrieve the points in the grid cells where all rotated and shifted versions of $q$ fall into. The runtime is then $O(e^{d/c}+k_c)=O(C_q(d,c)+k_c)$ ignoring the constant for matrix rotation that depends on $d$. Note that for Random Grids implementation we have $C_I(d,c)=C_q(d,c)$.  In contrast to $k$ in CANN-RS, $k_c$ here refers to the number of \emph{distinct colors} found in the enumerated cells.
As in \cite{aiger2013random}, the probability of success can be amplified to $1-\gamma$ by repeating the randomized indexing $\ln(1/\gamma)$ times, which increases the data structure, space and query time accordingly. The number of grids that we need in practice is much smaller than the above worst case depending on the intrinsic dimension of the data~\cite{aiger2013random}.

% In practice however, for $P$ with a small doubling dimension $\delta<\sqrt{d}$, the number of random grids can be reduced considerably to $O((ed/\delta)^\delta)$ and for $\delta\geq\sqrt{d}$ it would be $O(e^{\sqrt{d}}\sqrt{d}^\delta)$\cite{aiger2014reporting}. This is a worst case upper bound. We actually learn an appropriate value of $L$ from the data in the tuning phase (see below) and we found that it is much smaller in practice.

\paragraph{Constructing and querying CANN-RG} The above algorithms allow indexing colors of $P$ for a given $R$ such that for any query point $q$, the colors that have points at distance at most $R$ from $q$ are reported quickly. Given that we omitted the computation of point distances to enable efficient queries, we're still missing a way to compute the scores in Equation \ref{eq:metric}. We now show how we move from fixed radius Range Search to 1-NN. 

To fill this gap, let $r$ be a constant denoting the minimum distance between points in $P$ that we aim to distinguish. For each $l \in \{rc^0, rc^1,..., R\}$, we generate a sequence of random grid indexes $B^l = \{B^l_i, . . . , B^l_n\}$ of radius $l$. Then, given query point $q$, we query $q$ in all $B_i$ in order and keep only the closest (first observed) color.
This maps the list of colors to the $B^l_i$ they came from and thus to a $c$-approximate distance from the query.
Given these minimum distances, Equation \ref{eq:metric} provides a score per point and thus a ranking of all index images by  summing over all $q_j \in Q$.
This scoring operation increases the runtime of the query by logarithmic factor of $R/r$.
Note that CANN-RG is output sensitive on $k_c$, the number of actual neighbor colors we find for each query.

\section{Experiments}
\label{sec:experiments}
\subsection{Experimental setup}

\paragraph{Datasets:}
We evaluated our method on four public datasets from \cite{humenberger2022investigating}, ``Baidu-Mall",``Gangnam Station",``RobotCar Seasons" and ``Aachen Day-Night v1.1".
These datasets demonstrate performance in ``regular" outdoor scenarios as well as repetitive indoor environments. ``RobotCar Seasons" and ``Aachen Day-Night v1.1" have day and night subsets.

\paragraph{Metrics:}
We evaluated two metrics:
(1) The image retrieval performance using the same equal weighted barycenter (EWB) interpolation as in \cite{humenberger2022investigating} which is based solely on the retrieved images and their known poses.
(2) The effect on final localization quality using the existing localization pipeline from \cite{humenberger2022investigating} where camera localization is computed using only features from the top-k ranking images.
%The local features we used for retrieval are HOW, which were not previously benchmarked on localization and the r2d2 features which are commonly used for localization.
%These two metrics were also used in \cite{humenberger2022investigating} and we followed exactly the same process, replacing the pairfile (a list of retrieved images for each query image) by the one we computed using our method, for HOW and R2D2 respectively.

\paragraph{Local and global feature baselines:}
Following \cite{humenberger2022investigating,benchmarking_ir3DV2020}, we compared our method against state-of-the-art global features AP-GeM \cite{revaud2019learning}, DELG \cite{cao2020unifying}, DenseVLAD \cite{torii201524}, NetVLAD \cite{arandjelovic2016netvlad}. For local-features we compare performance and cost for both indexing and query to ASMK~\cite{TAJ13} with HOW and FIRE local features. Results for the latter were not previously published and only recently made available on the codebase for image retrieval methods. R2D2 features were computed using code from the same codebase. Storage cost for the baselines is discussed analytically.

\paragraph{Local feature types:}
We experiment with three state-of-the-art local image features: HOW~\cite{tolias2020learning}, FIRE~\cite{superfeatures} and R2D2 \cite{revaud2019r2d2}. These three approaches have different operation characteristics and thus show the power of CANN in being adaptable to different local features.
HOW and FIRE are designed for image retrieval, and are not suitable to the local feature matching part of the visual localization pipeline.
R2D2, on the other hand, is designed for image matching tasks and a common choice in structure-from-motion and visual localization evaluations \cite{jin2021image,humenberger2022investigating}.
We use a recent and lighter R2D2 version (referred to as ``Feather2d2 20k") described in \cite{humenberger2022investigating}'s codebase, where we can download the local features (the model is not publicly available). 
When using HOW and FIRE, our visual localization system requires indexing two different feature types: HOW for retrieval and R2D2 for matching.
When using R2D2, we only need to index one feature type -- which is appealing since it simplifies the overall system. For our experiments we used $1000$ per image for all indexed and query images and all methods.

\paragraph{Implementation details:}
We implemented CANN-RS and CANN-RG (Section \ref{sec:method}) in C++, given that it performs well for low intrinsic dimensions of the features: $32$D for R2D2 and $128$D for HOW. Even though CANN-RS can be implemented with any of-the-shelf range search data structures, we used Random Grids also here as it has the ability to exploit the fact that we know the range in advance.
The Random Grids were adjusted to different intrinsic dimensions by tuning its parameters, which is also required to trade off performance vs runtime using the $c$-approximation. Both our algorithms are very simple, trivially parallelized and are very fast (down to 20ms per query image). 

\paragraph{Tuning:}
The parameters of our metric are ${p}$ and ${R}$ and we tune them for each feature type separately. Note that in contrast to ASMK which creates a codebook that depends on the distribution of the data, CANN-RG and CANN-RS only tune for the metric itself.
One can therefore provide theoretic bounds of the (approximate) algorithmic result quality for a given metric. This may make CANN more resilient to different datasets which is not the case for codebook methods, even though the latter can perform better if the distribution of features between query and training set matches.
For CANN-RS, we set the grid cell size to slightly above $1/\sqrt{d}$ and the number of grids accordingly to balance result quality and runtime (see Section~\ref{CANN-RS}).
For CANN-RG we set $c=1.1$ in all datasets and the metric parameters ($p,R$) were tuned using a subset of $500$ queries from ``Baidu-Mall" separately per local feature type.
To the best of our knowledge, the datasets of~\cite{humenberger2022investigating} provide no tune/eval/test split and only the ``Baidu-Mall" has ground-truth available to enable tuning.
For ASMK we only evaluated R2D2 features, taking results for other features from ~\cite{humenberger2022investigating} or used previously unpublished results provided by the authors.
We train the ASMK codebook on ``GangnamStyle" as it is the largest set among the four.
To validate generalization, we used the same set of parameters for evaluation on all other datasets.

\subsection{Results}
As mentioned above, we evaluate the CANN-RG and CANN-RS algorithms on four large-scale datasets, in an outdoor, urban setting and covering an indoor scenario.
Following \cite{humenberger2022investigating,benchmarking_ir3DV2020} we evaluate across two regimes/metrics (``EWB" and ``SFM") discussed above. 
Figure~\ref{tab:all_results} shows our results of all methods and datasets with one figure per each metric.

In general, we can observe local features outperforming global features almost everywhere and by a large margin. Datasets that are more appropriate for global features are those that have many approximately similar viewpoints in the index so there is almost always one close neighbor for a given query image. Local features are naturally better where the query contains only partial overlap with the indexed images. 
Qualitative results are available in the appendix.

\paragraph{Runtime} One of the main advantages of CANN-RG (and CANN-RS as well) comparing to ASMK for image retrieval using local features is its simplicity and its runtime in both indexing and query. Table~\ref{tab:runtime} shows numbers across datasets using HOW features.
Since our implementation of CANN-RG and CANN-RS does not use GPU, we compared runtime on CPU using 48 cores.
The table does not contain the codebook creation for ASMK and tuning for CANN-RG. CANN-RG has a nice run-time/quality trade-off: In its upper bound quality, we have the results of CANN-RS and with CANN-RG can pay in quality for much better runtime. The significance of this is that CANN-RG can achieve runtimes of a few milliseconds for query image, which is otherwise only possible with global features. 
% Although it is inferior to the best possible results (with slower setup) it can still be better than global features.
Table~\ref{tab:runtime} provides results demonstrating the trade-off of runtime and quality.
To obtain a cheaper, yet representative quality measure, we compute the EWB using the top-1 retrieved image. The indexing time for CANN-RG is larger due to the fact that we have factor $O(\log{R})$ more data structures.

\setlength\tabcolsep{5pt}

\begin{table}[ht]
\begin{tabular}{|l|ccc|}
 \hline
  \multicolumn{1}{|l|}{}&\multicolumn{3}{c|}{Index} \\ 
  \hline
 Dataset & CANN-RS & CANN-RG & ASMK \\ 
 \hline
 Baidu    & 0.88  & 9.08    & 253.34   \\  
 Gangnam  & 6.41  & 168.49  & 1467.17  \\  
 Aachen   & 11.19 & 244.09  & 2782.43  \\  
 Robotcar & 33.02 & 852.12  & 8104.98  \\  
 \hline
  \multicolumn{1}{|l|}{}&\multicolumn{3}{c|}{Query} \\ 
  \hline
Baidu    & 0.37(12.47) & 0.02(12.12)  & 0.47(12.52) \\  
 Gangnam  & 1.6(12.66)  & 0.05(11.35)  & 0.41(11.03) \\  
 Aachen   & 1.38(29.1)  & 0.06(28.8)   & 0.48(28.0) \\  
 Robotcar & 5.29(94.2)  & 0.04(93.6)   & 0.53(91.0) \\ 
   \hline
\end{tabular}
\caption{Indexing and average runtime per query image (seconds) for CANN-RS, CANN-RG and ASMK using HOW features. An indication of quality/runtime trade-off can be taken from the simplified EWB metric,  computed using the top-1 retrieved image and provided in parentheses.}
\label{tab:runtime}
\end{table}
\vspace{-20pt}
\paragraph{Preliminary results on general image retrieval} To re-emphasize the generalization of the algorithm and it's scalability (20-50ms per query image), we also evaluated it for general image retrieval on the ROxford dataset. Global retrieval benchmarks evaluate the full rank of all indexed images, which requires also scoring the tail of the retrieved images.
Since ranking the tail of the index is not typically meaningful for local features, we evaluated a combination of CANN with global features by computing a weighted average of DELG and CANN-RG+HOW or CANN-RG+FIRE, for all image scores. We compare CANN and this combined approach to the SOTA for global/local features.
Very recently, a new method called Correlation Verification~\cite{DBLP:conf/cvpr/LeeS0K22} was published which is, to our knowledge the best performing method on the ROxford dataset. Correlation Verification however includes (significantly expensive) spatial verification of local features and is thus not comparable to CANN-RG which doesn't use geometry or spatial reasoning of features (out of the cameras).
Like for localization, spatial reasoning is an additional step that can be applied on top of CANN-RG. Table~\ref{tab:oxford} shows comparisons of SOTA approaches including~\cite{DBLP:conf/cvpr/LeeS0K22} with our proposed approach (bold).

\setlength\tabcolsep{3pt}

\begin{table}[H]
\small
%\begin{center}
\begin{tabular}{lcc}
\hline
(A) Local feature aggregation&Medium&Hard\\
%DELF-ASMK*+SP~\cite{Hyeonwoo2017large,Radenovic2018benchmarking} &67.8 &43.1 \\
DELF-D2R-R-ASMK* (GLDv1)~\cite{teichmann2019detect} &73.3 & 47.6\\
%+ SP (Rerank Top-100))~\cite{teichmann2019detect} &76.0 &52.4\\
R50-HOW-ASMK,n=2000~\cite{Tolias2019learning} &79.4 &56.9\\
\hline
(B) Global feature\\
R101-GeM~\cite{Radenovic2018Finetuning,DBLP:conf/cvpr/SimeoniAC19}& 65.3 &39.6\\
%+DSM (Rerank Top-100)~\cite{DBLP:conf/cvpr/SimeoniAC19}& 65.3 &39.2\\
R101-GeM-AP (GLDv1)~\cite{DBLP:conf/iccv/RevaudARS19}& 66.3 &42.5\\
R101-GeM+SOLAR (GLDv1)~\cite{DBLP:conf/eccv/NgBTM20}&69.9 &47.9\\
R50-DELG (Global-only, GLDv2-clean)~\cite{DBLP:conf/eccv/CaoAS20}&73.6 &51.0\\
%+ GV (Rerank Top-100~\cite{DBLP:conf/eccv/CaoAS20}& 78.3 &57.9\\
%+ GV (Rerank Top-200)~\cite{DBLP:conf/eccv/CaoAS20,DBLP:conf/iccv/TanYO21}&79.2 &57.5\\
%+ RRT (Rerank Top-100)~\cite{DBLP:conf/iccv/TanYO21}& 78.1 &60.2\\
%+ RRT (Rerank Top-200)~\cite{DBLP:conf/iccv/TanYO21}&79.5 &62.5\\
R101-DELG (Global-only, GLDv2-clean)~\cite{DBLP:conf/eccv/CaoAS20}&76.3 &55.6\\
%+ GV (Rerank Top-100)~\cite{DBLP:conf/eccv/CaoAS20}& 81.2 &64.0\\
%+ RRT (Rerank Top-100)~\cite{DBLP:conf/eccv/CaoAS20}& 79.9 &64.1 \\
%+ SuperGlue (Rerank Top-100)~\cite{DBLP:conf/eccv/CaoAS20,DBLP:conf/cvpr/SarlinDMR20}&79.7 &62.1\\
R50-DOLG (GLDv2-clean)~\cite{DBLP:conf/iccv/YangHFSXLDH21}&80.5 &58.8\\
R101-DOLG (GLDv2-clean)~\cite{DBLP:conf/iccv/YangHFSXLDH21}&81.5 &61.1\\
R101-CVNet-Global (GLDv2-clean)~\cite{DBLP:conf/cvpr/LeeS0K22}&80.2 & 63.1\\
\hline
\textbf{DELG+CANN-FIRE (weighted)}&\textbf{82.4} &62.3\\
\textbf{DELG+CANN-HOW (weighted)}&\textbf{83.3} &\textbf{64.2}\\
%+\textbf{CVNet-Rerank (Rerank Top-400)}~\cite{DBLP:conf/cvpr/LeeS0K22}&\textbf{87.2} &\textbf{75.9}\\
\end{tabular}
%\end{center}
\caption{Results of DELG+CANN compared to state-of-the-art reranking (local aggregation and global features) on ROxford (numbers of related work from~\cite{DBLP:conf/cvpr/LeeS0K22})}.
    \label{tab:oxford}
\end{table}

\paragraph{Limitations.} Using local features throughout the stack requires that the entire map fit in memory. Approaches that use global features can be more easily scaled, in that the local features per spatial region are kept out-of-memory and are only loaded after image retrieval.

\begin{table*}[p]
\begin{tabular}{lll}
\hspace{-20pt}\includegraphics[width=7cm]{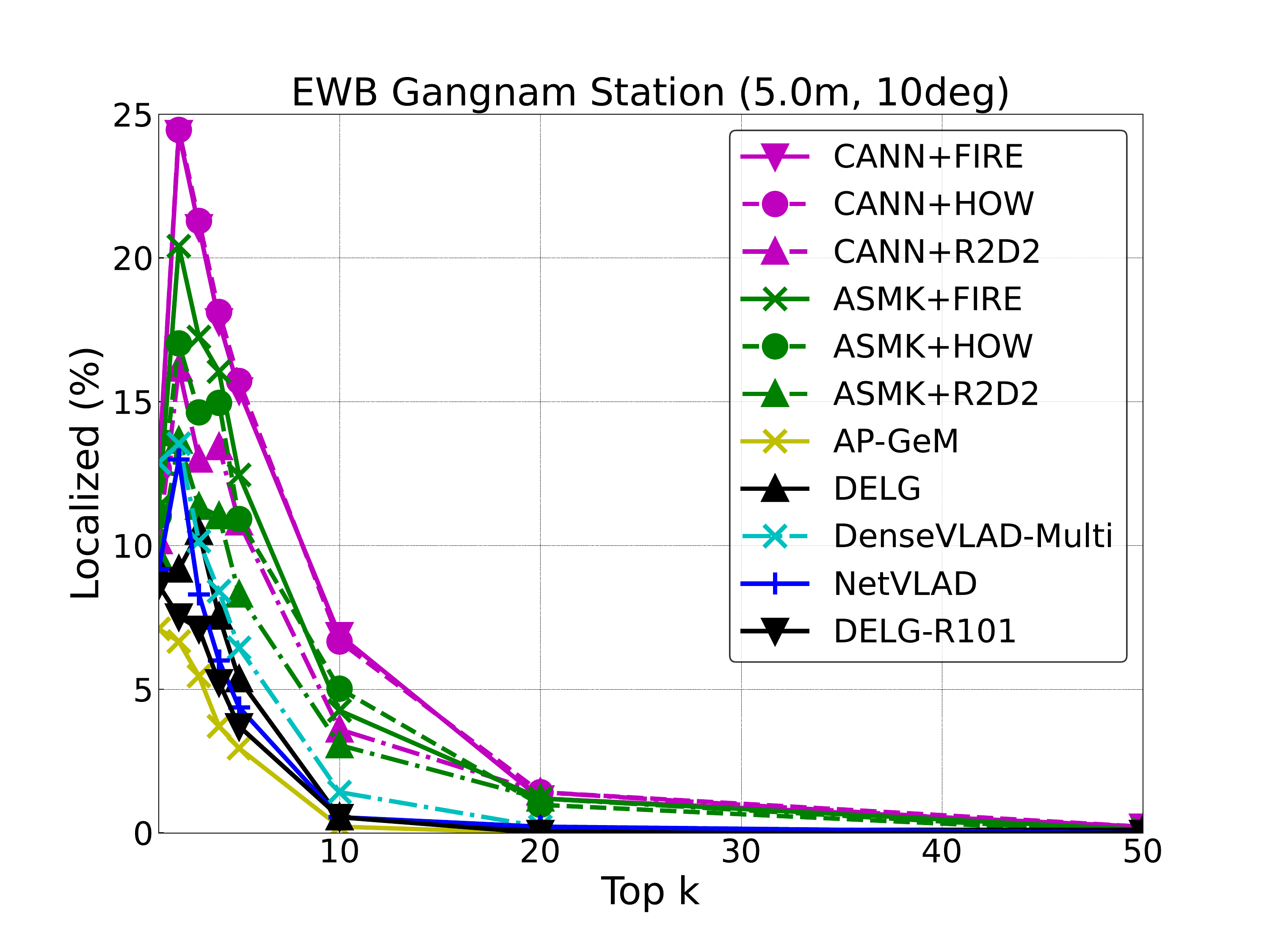} & \hspace{-28pt} \includegraphics[width=7cm]{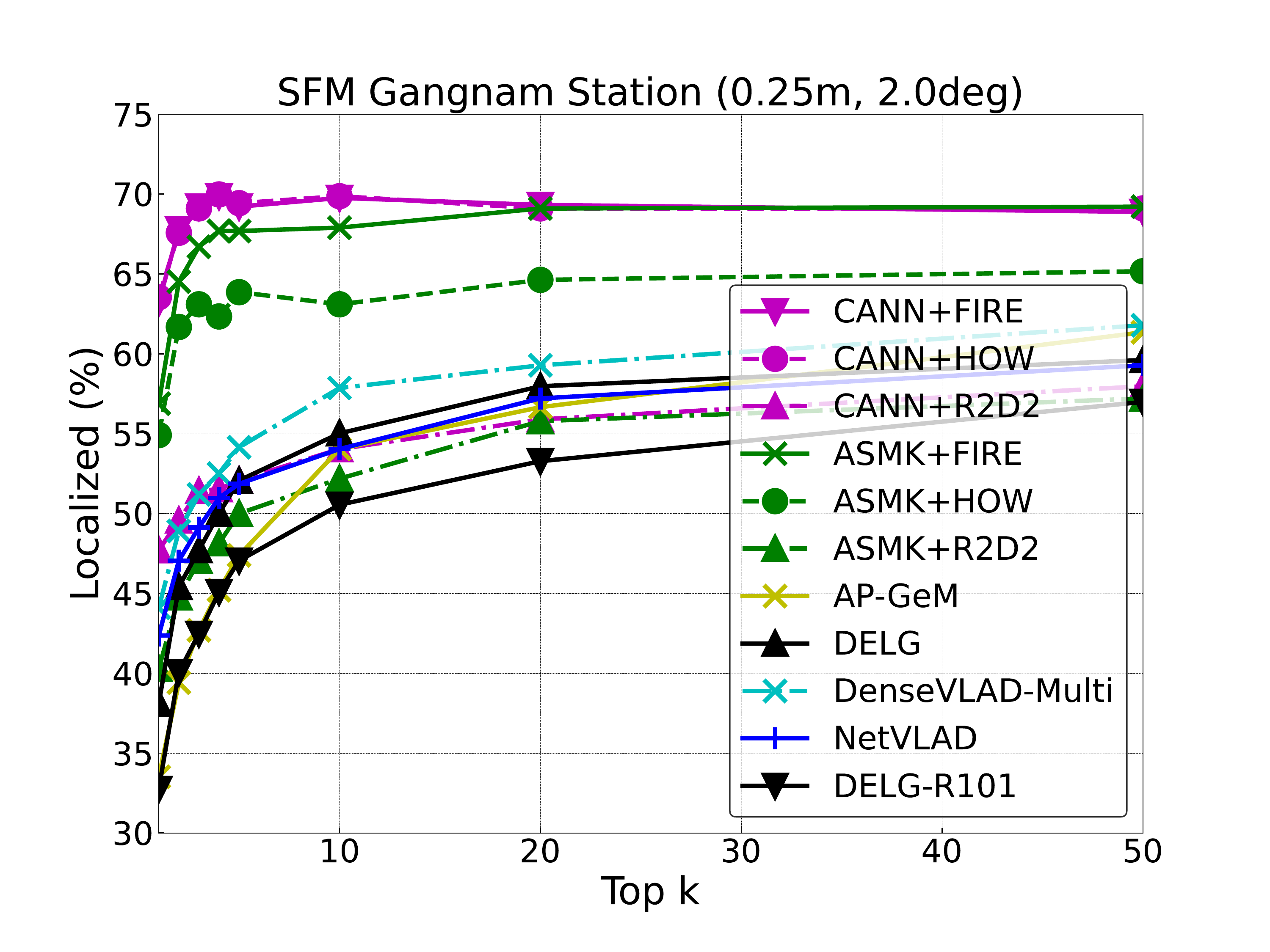} & \hspace{-28pt}
\includegraphics[width=7cm]{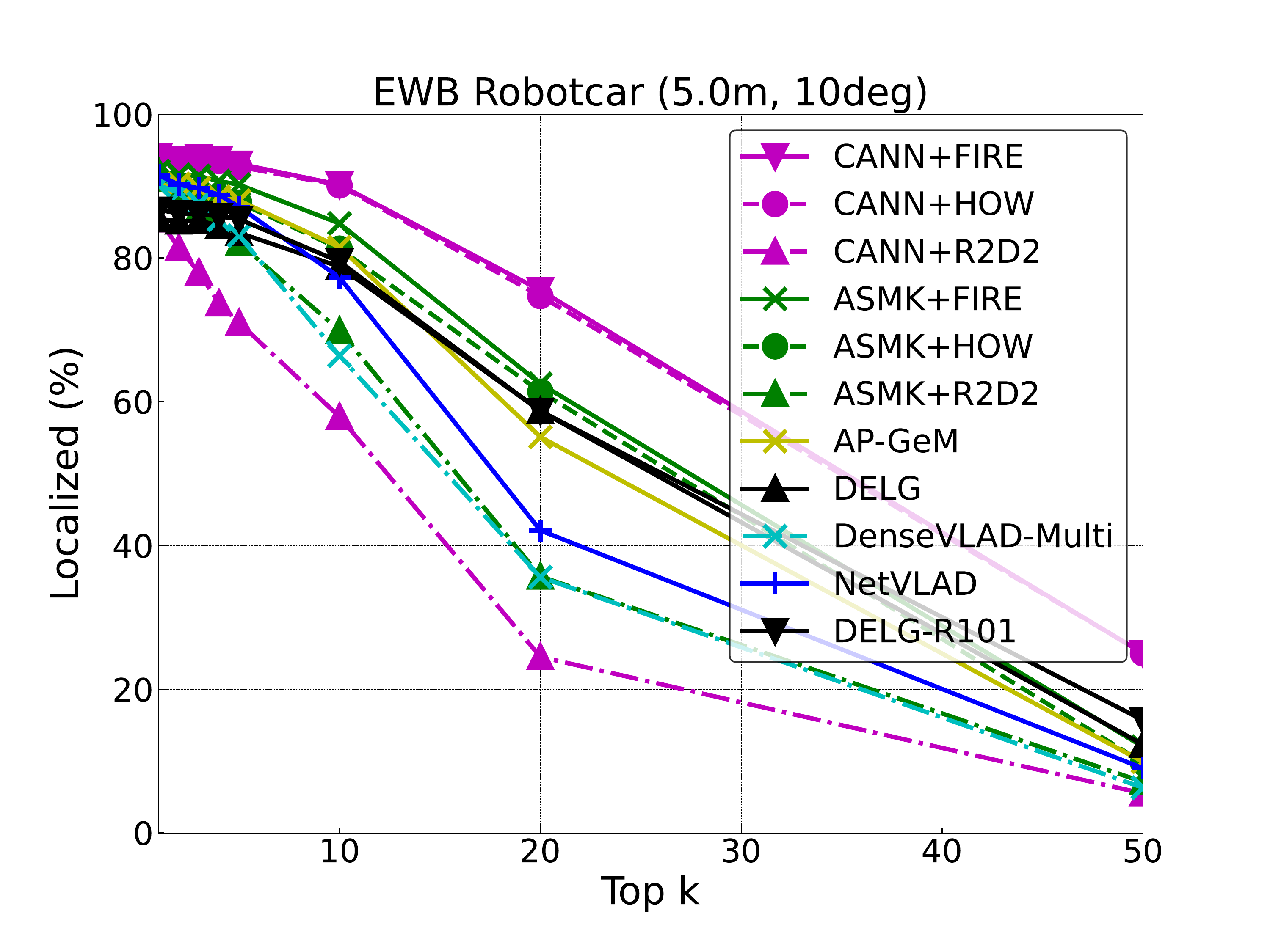} \\ \hspace{-20pt}\includegraphics[width=7cm]{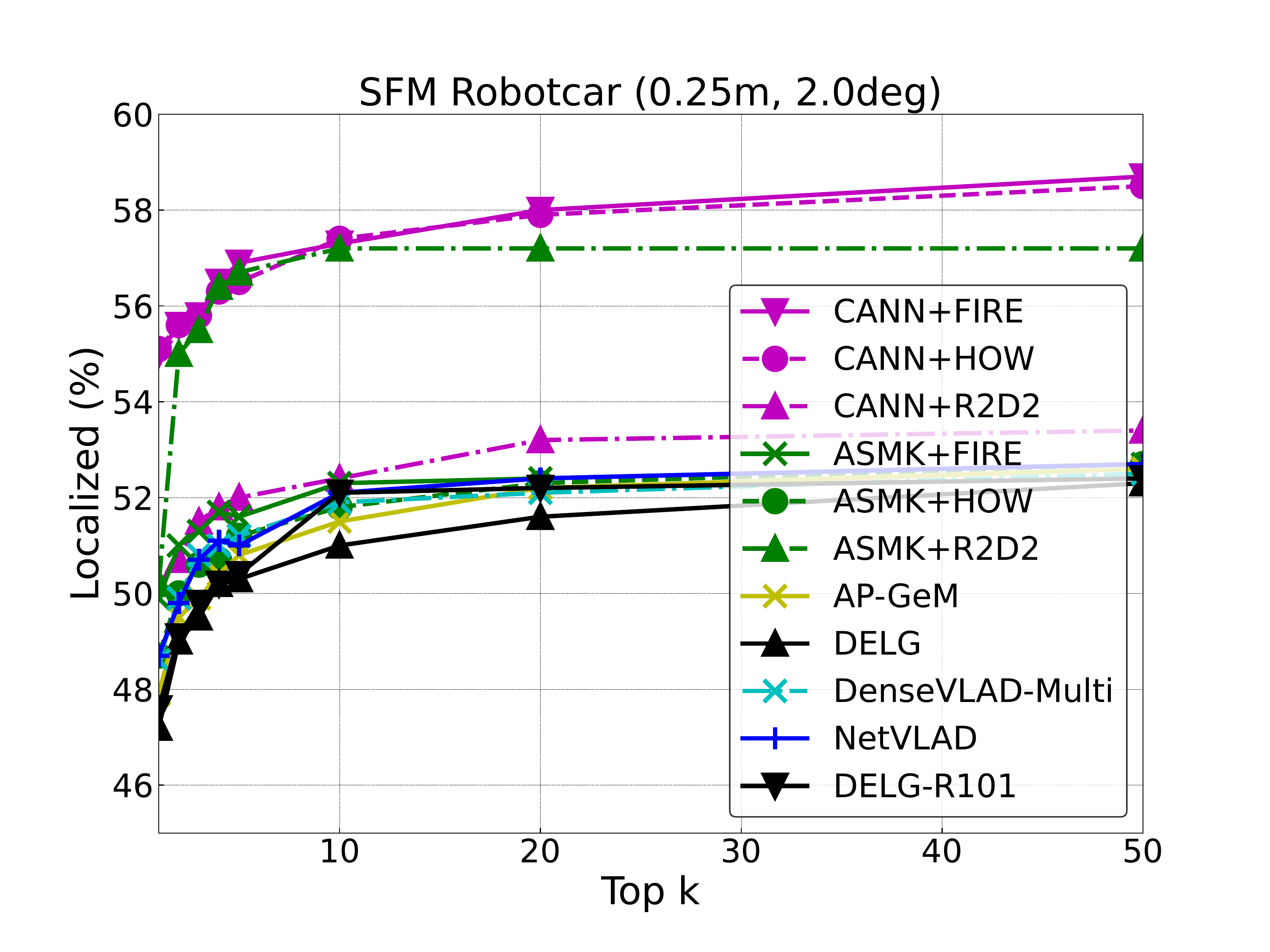} &\hspace{-28pt}
\includegraphics[width=7cm]{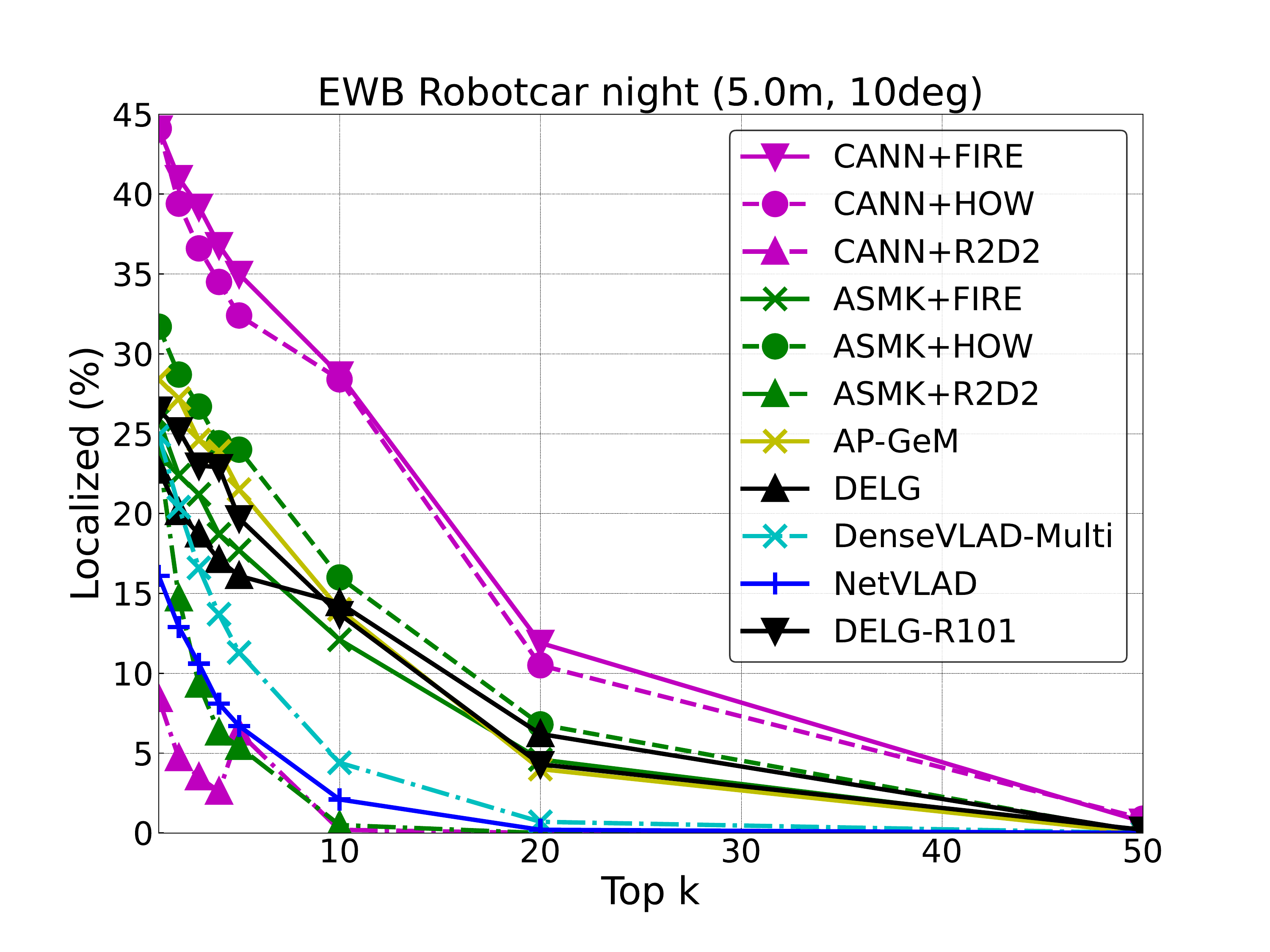} & \hspace{-28pt} \includegraphics[width=7cm]{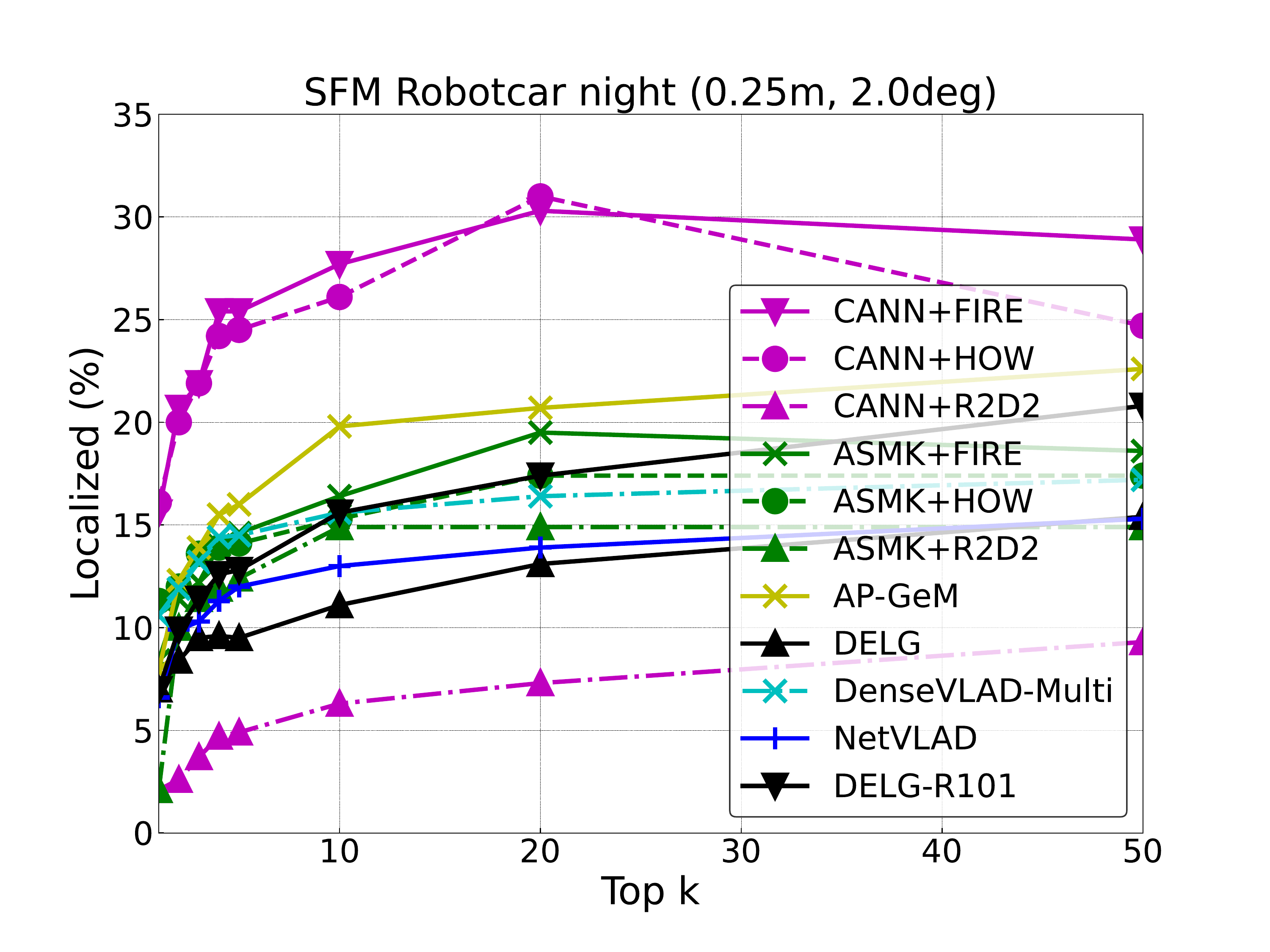} \\ 
\hspace{-20pt}\includegraphics[width=7cm]{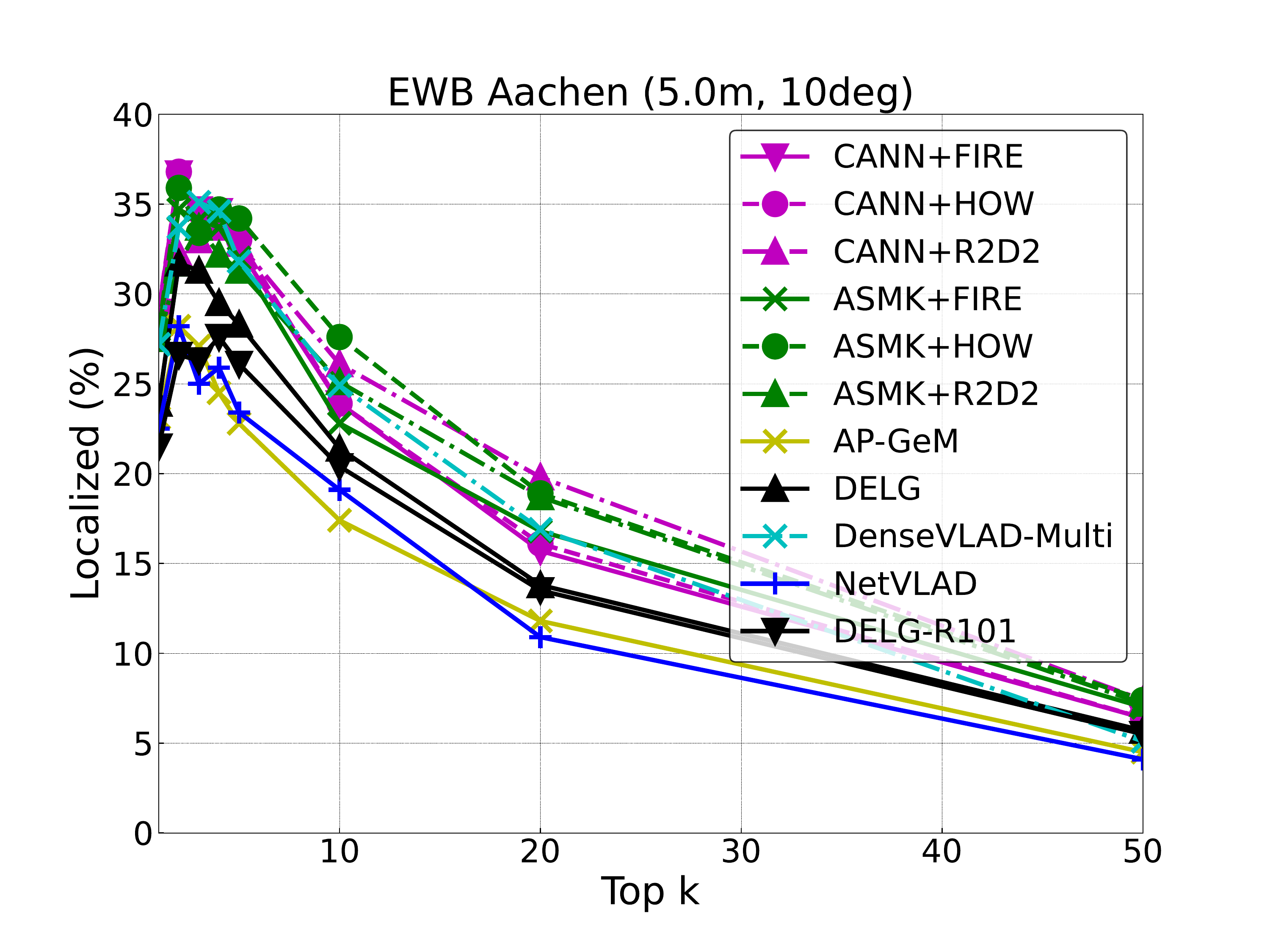} &\hspace{-28pt} \includegraphics[width=7cm]{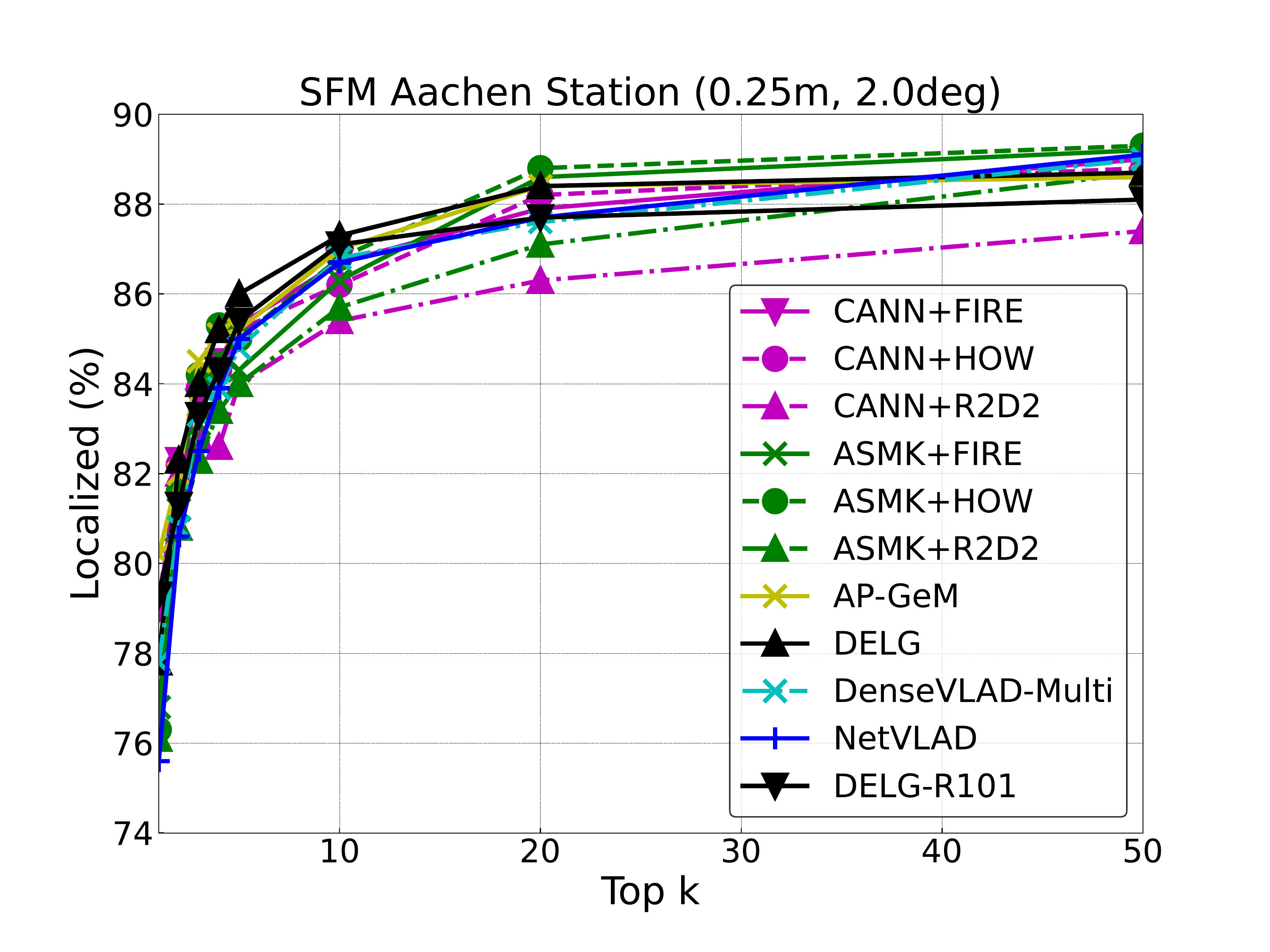} &\hspace{-28pt}
\includegraphics[width=7cm]{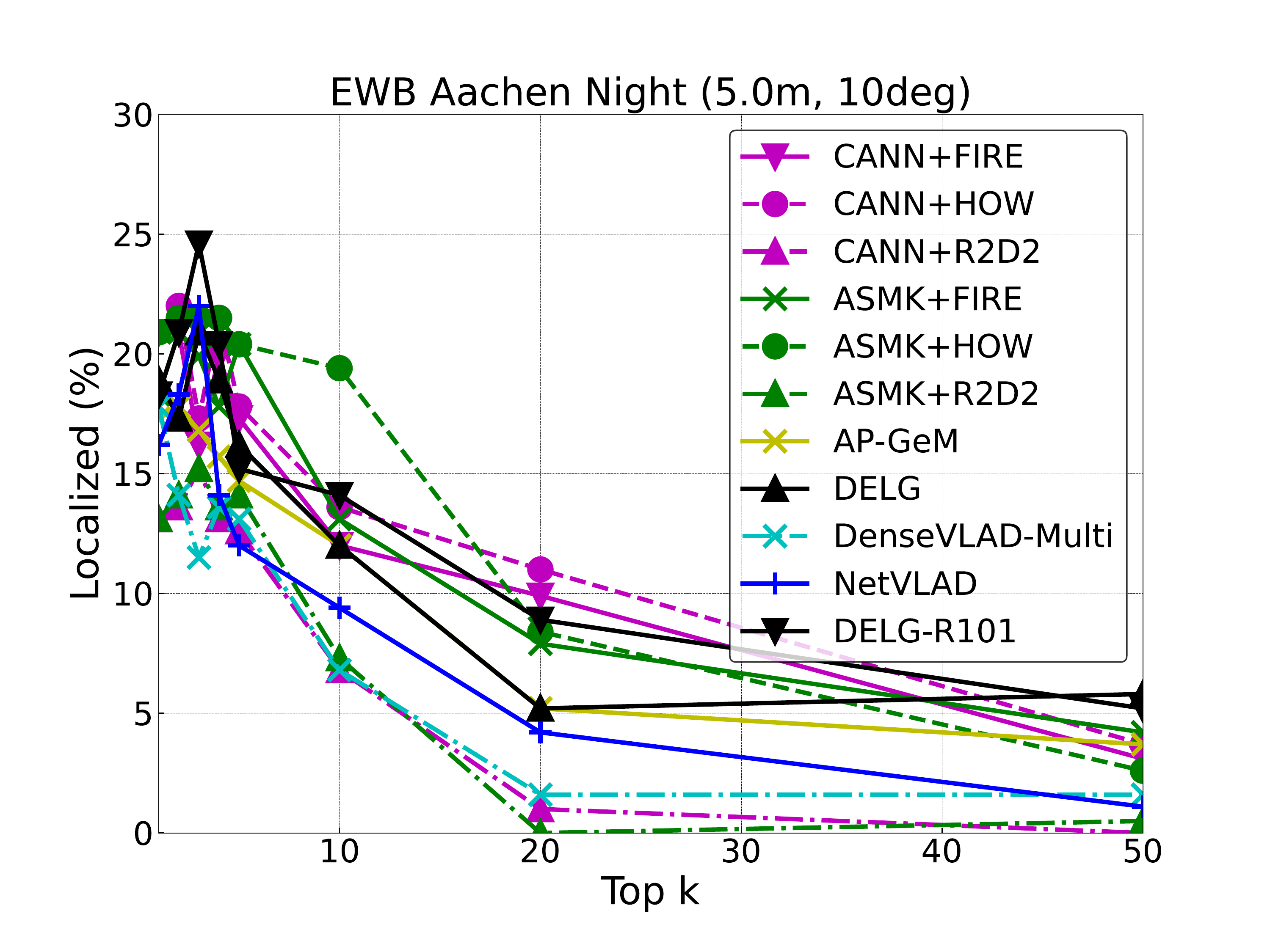} \\ \hspace{-20pt}\includegraphics[width=7cm]{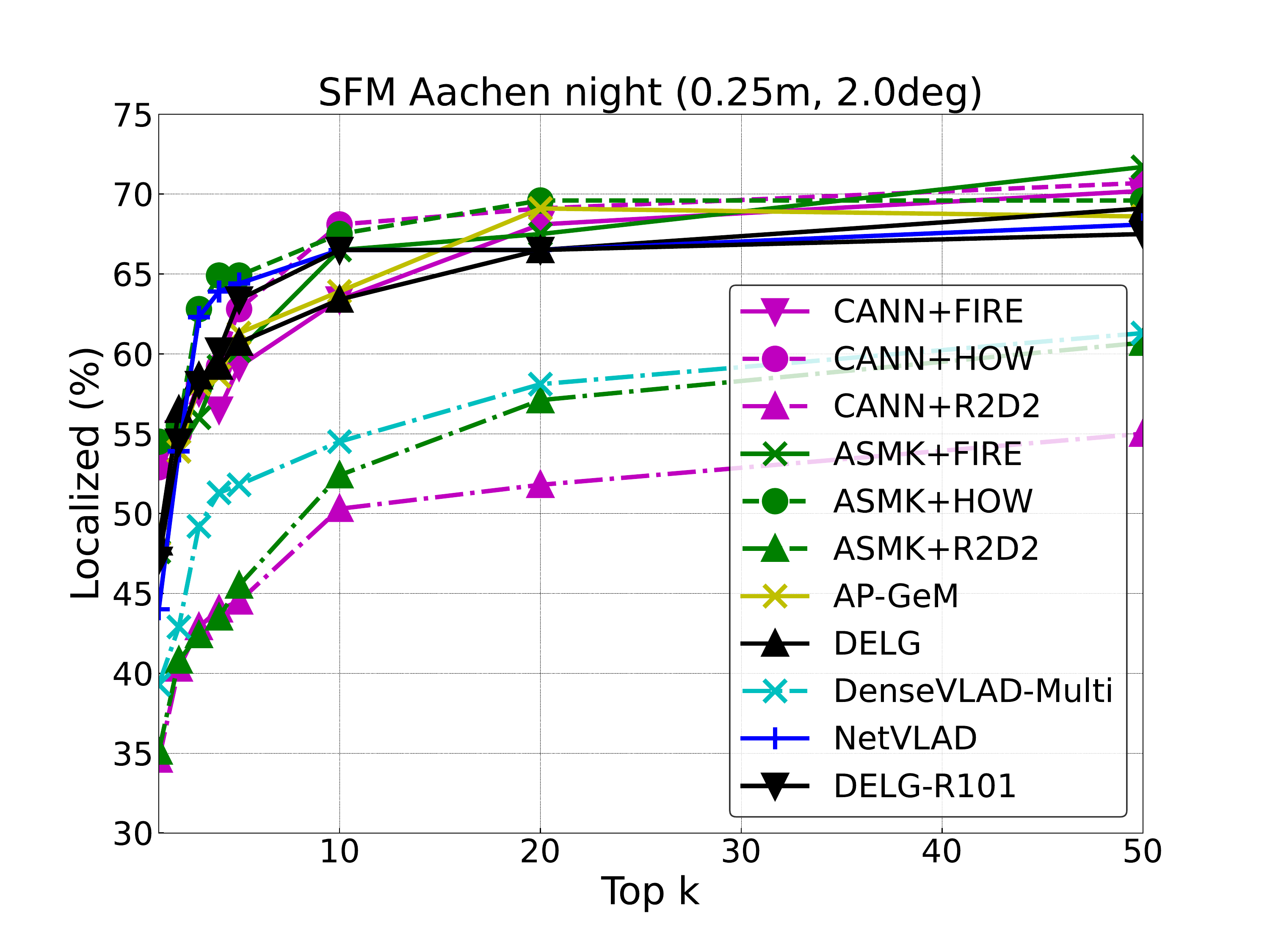} &\hspace{-28pt}
\includegraphics[width=7cm]{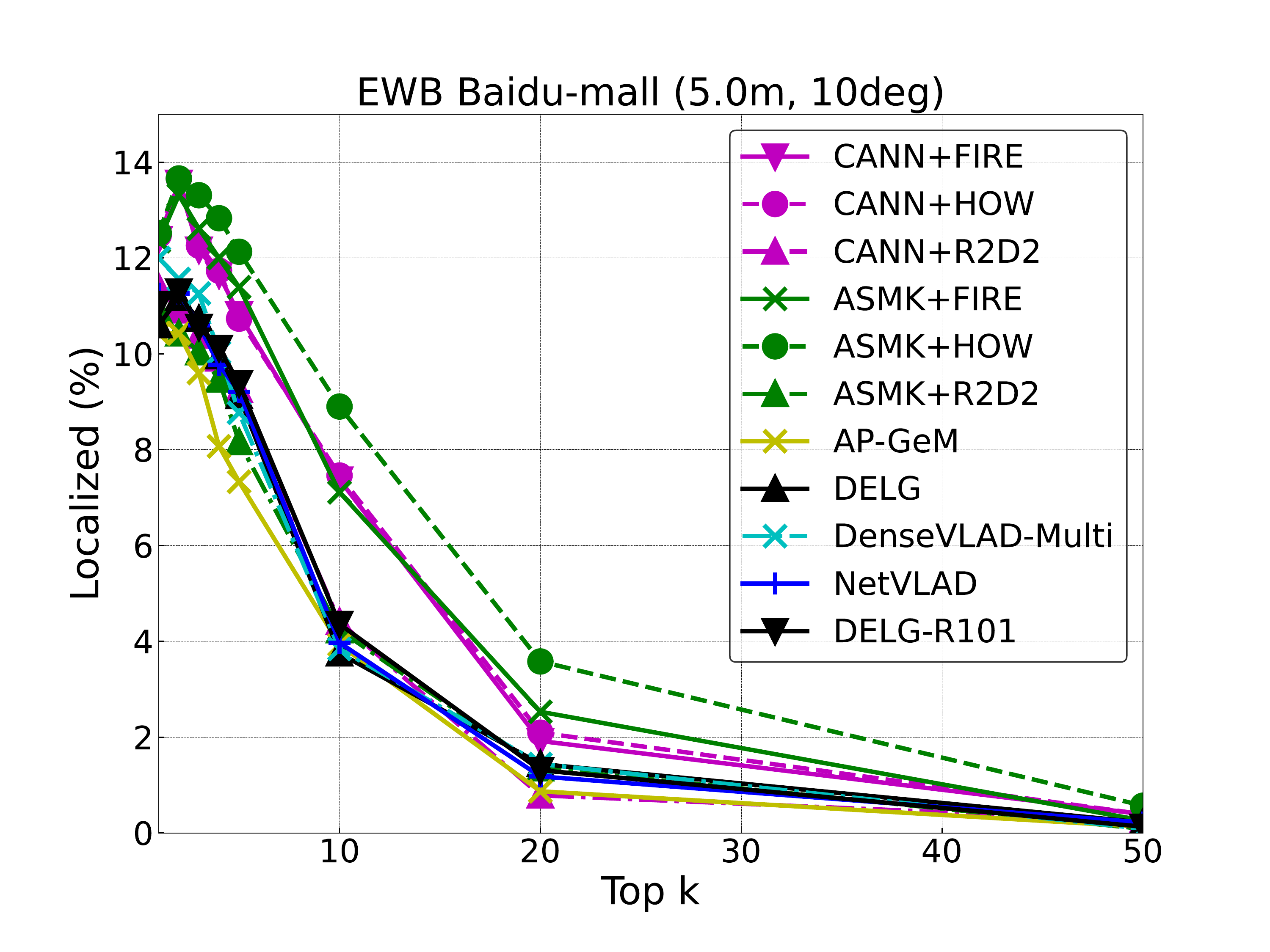} &\hspace{-28pt} \includegraphics[width=7cm]{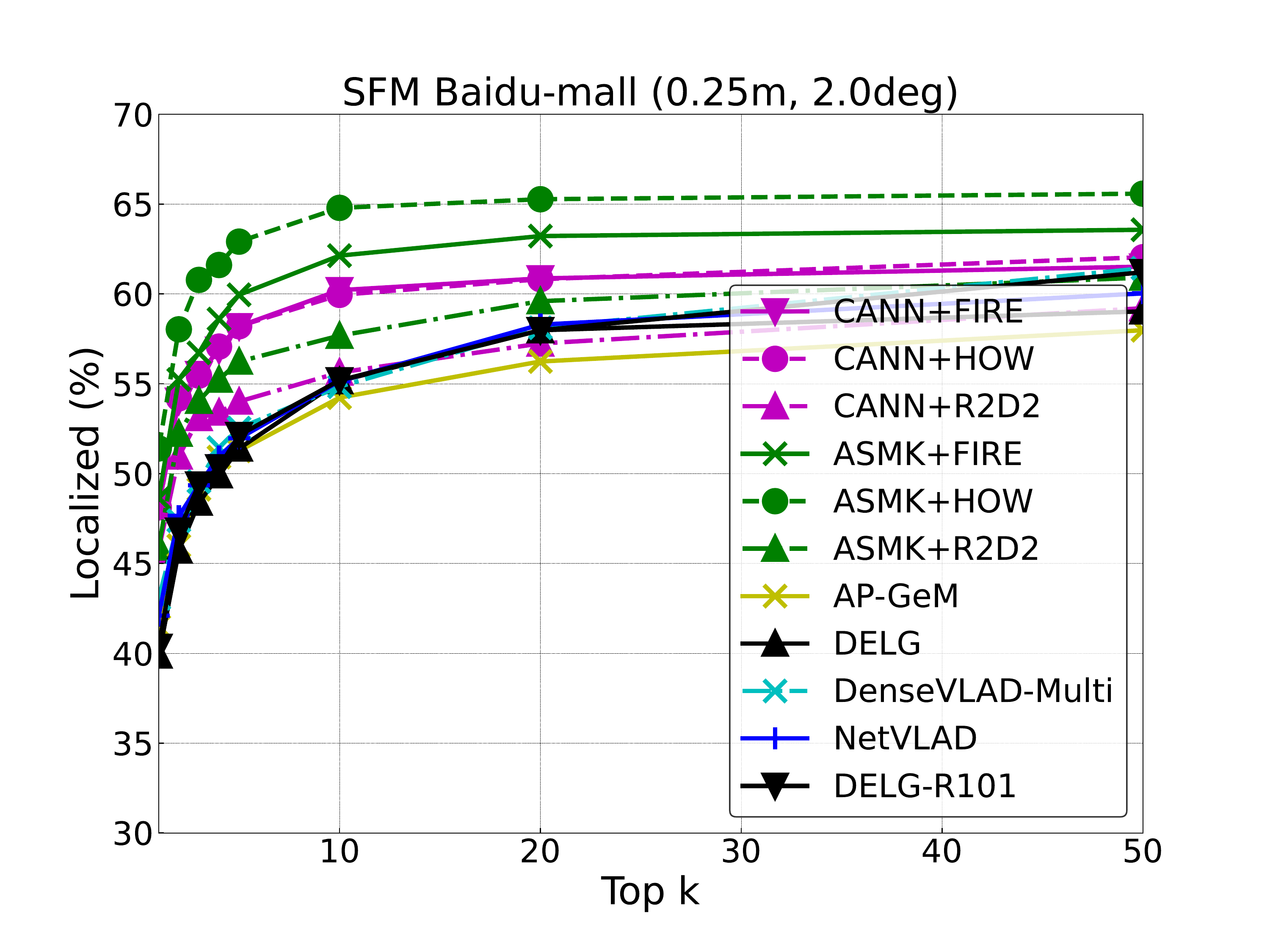} \\ 

\end{tabular}
\caption{Results from four public benchmarks. Following \cite{humenberger2022investigating,benchmarking_ir3DV2020} we evaluate across image retrieval using equal weighted barycenter (EWB) interpolation and final localization quality using the existing localization pipeline from \cite{humenberger2022investigating} [SFM].}
\label{tab:all_results}
\end{table*} 

% \begin{figure*}[ht]
% \begin{center}
%       \includegraphics[width=\textwidth]{images/qualitative_1.pdf}
% \end{center}
% \vspace{-10pt}
%   \caption{Qualitative results for a challenging query from the Baidu-Mall dataset, where we compare CANN-RG (bottom row) against state of the art global features. The query image is on the left and the top 5 retrieved images are on the right. Our method retrieves all correct images, while other methods retrieve at least 2 incorrect images among the top 5.
%   Situations like this, were the top database images have strong occlusions are particularly difficult to solve correctly with global features and demonstrate the benefit of local feature based approaches.}
% \label{fig:qualitative}
% \end{figure*}

% \begin{figure*}[ht]
% \begin{center}
%       \includegraphics[width=\textwidth]{images/nn1.png} \\
%       \includegraphics[width=\textwidth]{images/nn2.png} \\
%       \includegraphics[width=\textwidth]{images/nn3.png} \\
%       \includegraphics[width=\textwidth]{images/nn4.png} \\
%       \includegraphics[width=\textwidth]{images/nn5.png} \\
%       \includegraphics[width=\textwidth]{images/hard_nn.png} \\
% \end{center}
% \vspace{-10pt}
%   \caption{Qualitative results from RG-CANN. The query image is on the left and the top 1-5 images retrieved are in the right. We found the method to perform well, also for harder cases like the one in the last row.}
% \label{fig:qualitative}
% \end{figure*}
\section{Conclusions}
\label{sec:conclusions}

In this paper, we proposed CANN, a novel nearest neighbor searching approach that finds the best matches in both appearance and geometry space to improve visual localization using only local features.
Unlike the state-of-the-art in the field, which uses global features for image retrieval and local features for 2D-3D matching, our approach uses only local features, while providing significantly better performance than the state-of-the-art at very competitive runtime cost.
By providing the relevant metric and theoretical foundation of the algorithm, as well as two efficient algorithmic solutions, we hope to inspire a revived interest in solving visual localization with local features only.

\clearpage % force a pagebreak and flush all deferred `table` and `figure` environments

\appendix

\section{Additional Qualitative Results}
\label{sec:sec1}

\begin{figure*}[ht]
\begin{center}
      \includegraphics[width=0.8\textwidth]{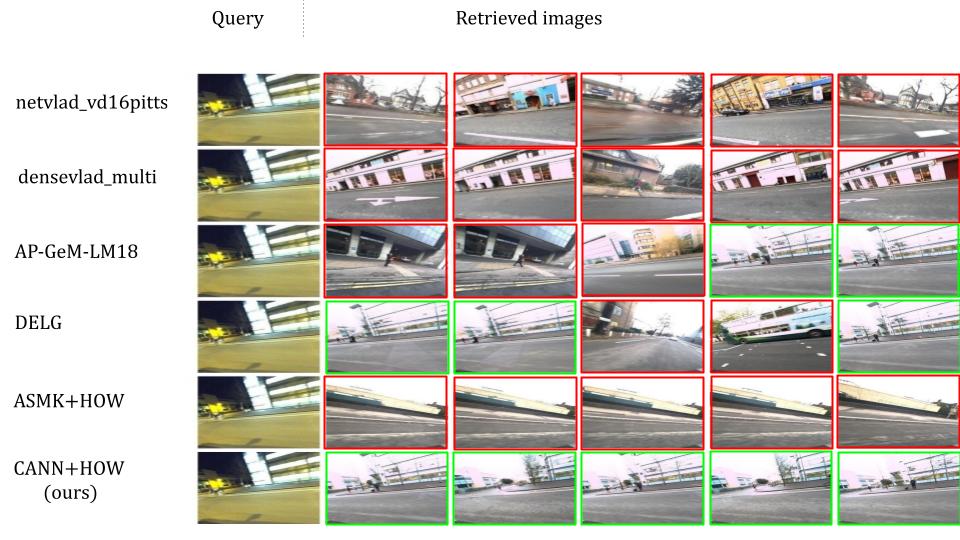}
      \includegraphics[width=0.8\textwidth]{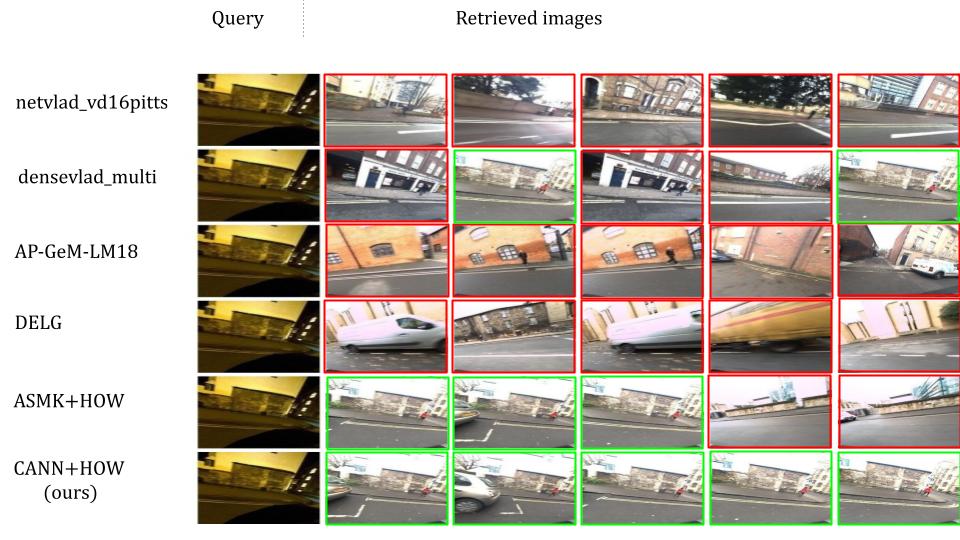}
\end{center}
\vspace{-10pt}
  \caption{Robotcar}
\label{fig:supp_aachen_1}
\end{figure*}

\begin{figure*}[ht]
\begin{center}
      \includegraphics[width=0.8\textwidth]{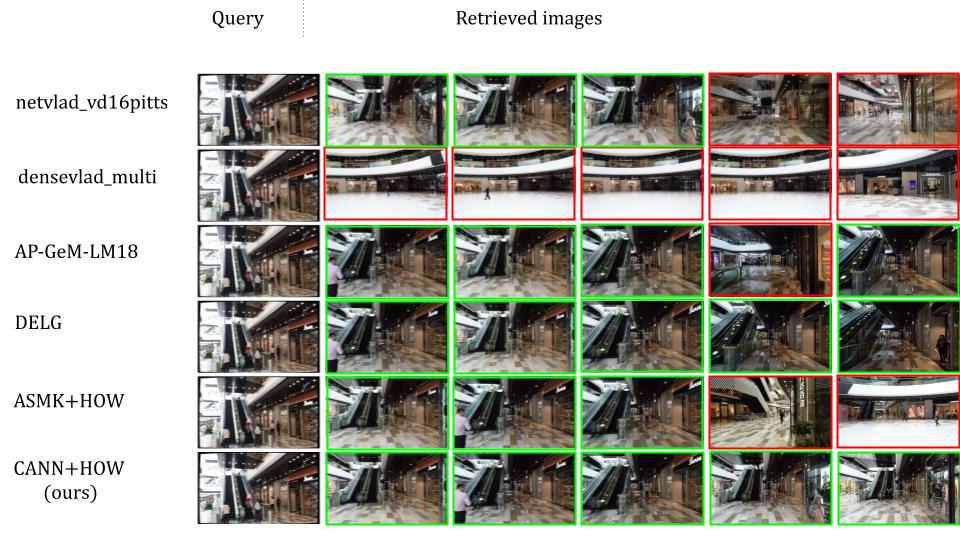}
      \includegraphics[width=0.8\textwidth]{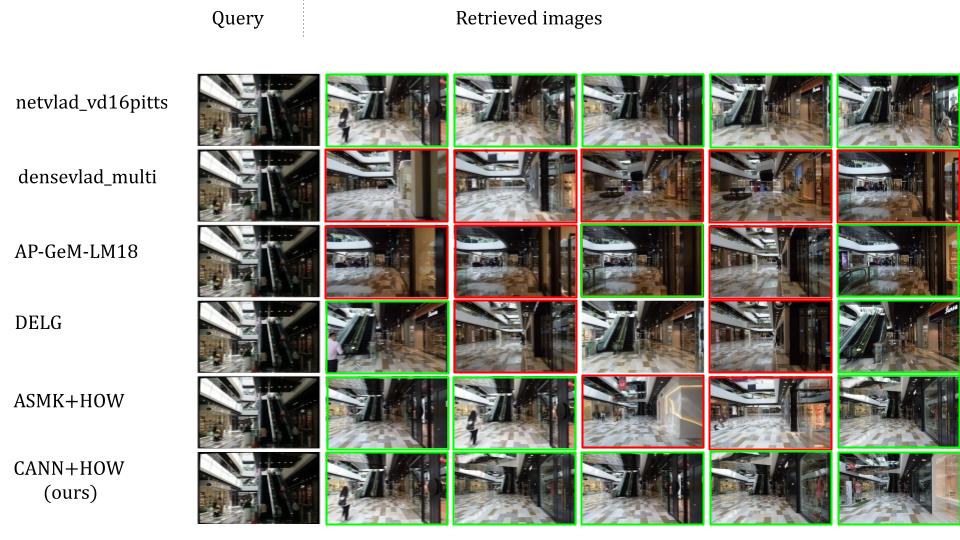}
\end{center}
\vspace{-10pt}
  \caption{Gangnam}
\label{fig:supp_aachen_2}
\end{figure*}

\begin{figure*}[ht]
\begin{center}
      \includegraphics[width=0.8\textwidth]{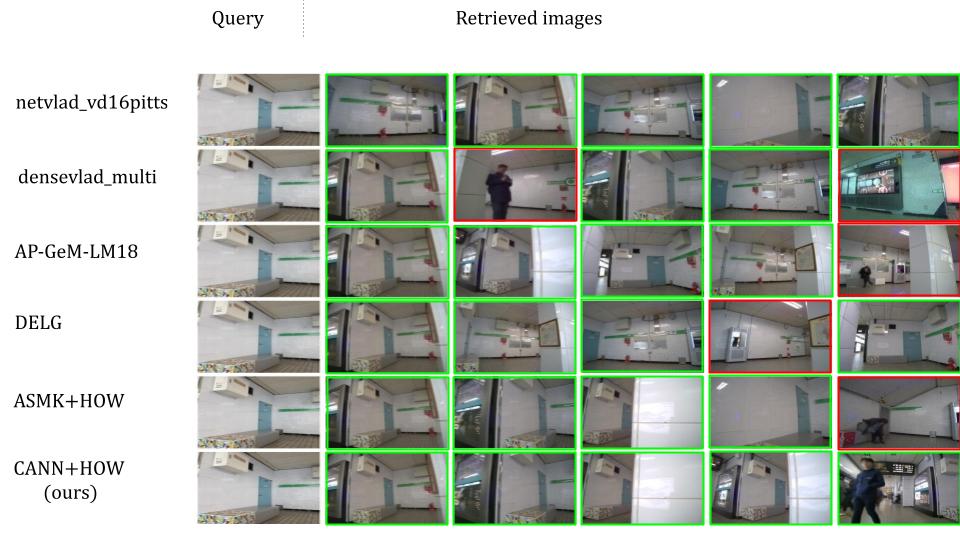}
      \includegraphics[width=0.8\textwidth]{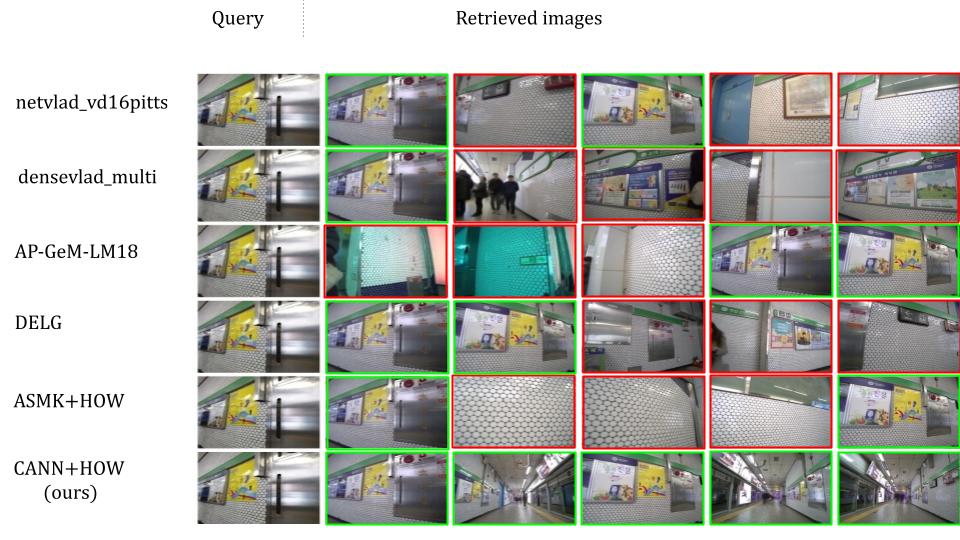}
\end{center}
\vspace{-10pt}
  \caption{Baidu}
\label{fig:supp_aachen_3}
\end{figure*}

\begin{figure*}[ht]
\begin{center}
      \includegraphics[width=0.8\textwidth]{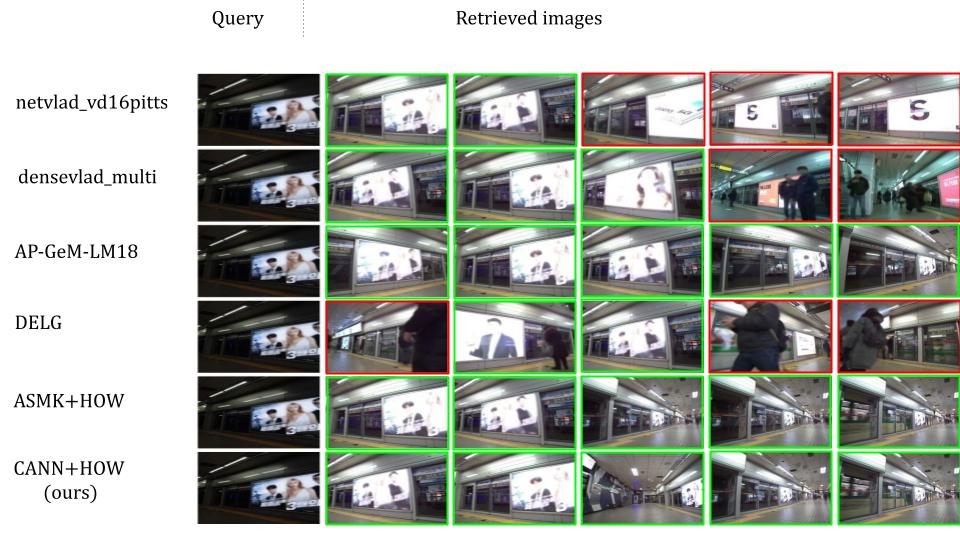}
\end{center}
\vspace{-10pt}
  \caption{Baidu}
\label{fig:supp_aachen_4}
\end{figure*}

\begin{figure*}[ht]
\begin{center}
      \includegraphics[width=0.8\textwidth]{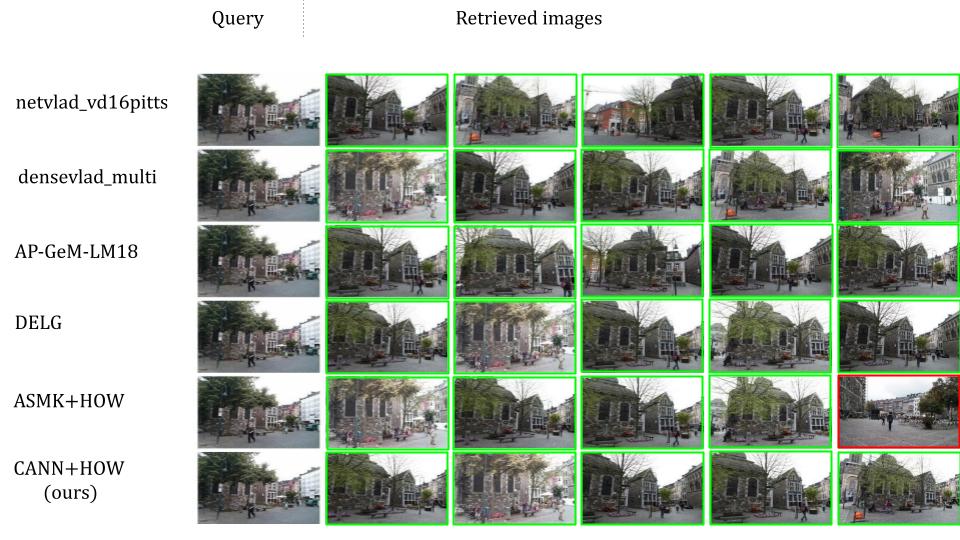}
\end{center}
\vspace{-10pt}
  \caption{Aachen}
\label{fig:supp_aachen_5}
\end{figure*}

\begin{figure*}[ht]
\begin{center}
      \includegraphics[width=0.8\textwidth]{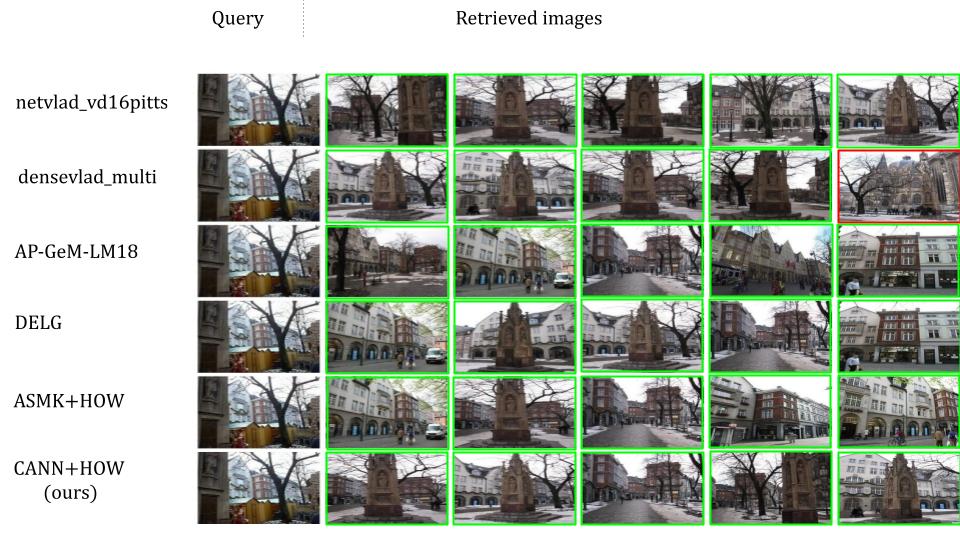}
\end{center}
\vspace{-10pt}
  \caption{Aachen}
\label{fig:supp_aachen_6}
\end{figure*}

We include additional qualitative results in Figures \ref{fig:supp_aachen_1},\ref{fig:supp_aachen_2},\ref{fig:supp_aachen_3},\ref{fig:supp_aachen_4},\ref{fig:supp_aachen_5},\ref{fig:supp_aachen_6} taken from all datasets, showing that CANN retrieves good results also in images with heavy occlusions. Cases like these, where there is only partial overlap between the query image and database images are very difficult for global features. We use HOW~\cite{tolias2020learning} for local features with both CANN-RG (ours) and ASMK~\cite{tolias2015image}. The query image is on the left and the top 5 retrieved images are on the right. Our method retrieves all correct images, while other methods retrieve occasionally incorrect images ranked high among the top 5. We see that some global methods retrieve incorrect images due to scene clutter or high-frequency textures, while CANN provides diverse set of correct results. In several cases, we see that CANN+HOW outperforms ASMK+HOW. Retrieved images are marked red (bad) or green (good).

\clearpage

%%%%%%%%% REFERENCES
{\small
\bibliographystyle{ieee_fullname}
\bibliography{literature/cann_iccv23}
}

\end{document}